\documentclass[lettersize,journal]{IEEEtran}
\usepackage{amsmath,amsfonts}
\usepackage{algorithmic}
\usepackage{array}
\usepackage[caption=false,font=normalsize,labelfont=sf,textfont=sf]{subfig}
\usepackage{textcomp}
\usepackage{stfloats}
\usepackage{url}
\usepackage{verbatim}
\usepackage{graphicx}

\usepackage{color}
\usepackage{booktabs}
\usepackage{multirow}
\usepackage{pifont} % DO NOT CHANGE THIS AND DO NOT ADD ANY OPTIONS TO IT
\usepackage{setspace}

\def\BibTeX{{\rm B\kern-.05em{\sc i\kern-.025em b}\kern-.08em
    T\kern-.1667em\lower.7ex\hbox{E}\kern-.125emX}}
\usepackage{balance}
\begin{document}
\title{Dynamic Modality-Camera Invariant Clustering for Unsupervised Visible-Infrared Person Re-identification}
\author{Yiming Yang, Weipeng Hu, Haifeng Hu
\thanks{Yiming Yang and Haifeng Hu are with the School of Electronics and Information Technology, Sun Yat-sen University, Guangzhou 510006, China. (e-mail: yangym53@mail2.sysu.edu.cn; huhaif@mail.sysu.edu.cn).\\
	
Weipeng Hu is with the School of Electrical and Electronic Engineering (EEE), Nanyang Technological University, Singapore 639798. (e-mail: weipeng.hu@ntu.edu.sg).}}

\markboth{Journal of \LaTeX\ Class Files,~Vol.~18, No.~9, September~2020}%
{How to Use the IEEEtran \LaTeX \ Templates}

\maketitle

\begin{abstract}
Unsupervised learning visible-infrared person re-identification (USL-VI-ReID) offers a more flexible and cost-effective alternative compared to supervised methods. This field has gained increasing attention due to its promising potential. Existing methods simply cluster modality-specific samples and employ strong association techniques to achieve instance-to-cluster or cluster-to-cluster cross-modality associations. However, they ignore cross-camera differences, leading to noticeable issues with excessive splitting of identities. Consequently, this undermines the accuracy and reliability of cross-modal associations. To address these issues, we propose a novel Dynamic Modality-Camera Invariant Clustering (DMIC) framework for USL-VI-ReID. Specifically, our DMIC naturally integrates Modality-Camera Invariant Expansion (MIE), Dynamic Neighborhood Clustering (DNC) and Hybrid Modality Contrastive Learning (HMCL) into a unified framework, which eliminates both the cross-modality and cross-camera discrepancies in clustering. MIE fuses inter-modal and inter-camera distance coding to bridge the gaps between modalities and cameras at the clustering level. DNC employs two dynamic search strategies to refine the network's optimization objective, transitioning from improving discriminability to enhancing cross-modal and cross-camera generalizability. Moreover, HMCL is designed to optimize instance-level and cluster-level distributions. Memories for intra-modality and inter-modality training are updated using randomly selected samples, facilitating real-time exploration of modality-invariant representations. Extensive experiments have demonstrated that our DMIC addresses the limitations present in current clustering approaches and achieve competitive performance, which significantly reduces the performance gap with supervised methods.

\end{abstract}

\begin{IEEEkeywords}
Person re-identification (Re-ID), cross-modality, unsupervised learning, clustering.
\end{IEEEkeywords}

\section{Introduction}

\IEEEPARstart{P}{erson} re-identification is employed to identify and locate specific individuals among pedestrians captured in multiple camera surveillance scenarios \cite{Ge2024Structured,wu2023unsupervised}. This technology have been significantly developed during the last decade due to its important role in the fields of multimedia data retrieval and criminal investigation \cite{ye2021deep,chen2017person,yu2018unsupervised}. Earlier works on person Re-ID focus on the retrieval of pedestrian images captured by RGB cameras. However, such methods falter in low-light conditions. This is primarily due to  the inherent limitation of RGB cameras in acquiring high-definition images in darkness. Therefore, visible-infrared person re-identification (VI-ReID) is proposed and employed to form 24-hour surveillance system, which aims to match infrared images under poor illumination with visible images under good illumination.

The current VI-ReID methods focus on generation and subspace mapping technique to learn modality-invariant representations, achieving notable success \cite{ye2020cross,lu2023tri,choi2020hi,wang2019learning}. However, their dependence on manually annotated associations between visible and infrared modalities can hinder the scalability and deployment of the VI-ReID model. Unsupervised Learning Visible-Infrared Person Re-Identification (USL-VI-ReID) is introduced to eliminate this reliance on annotations and gains increasing attention due to its promising potential.

\begin{figure}
	\centerline{\includegraphics[width=0.5\textwidth]{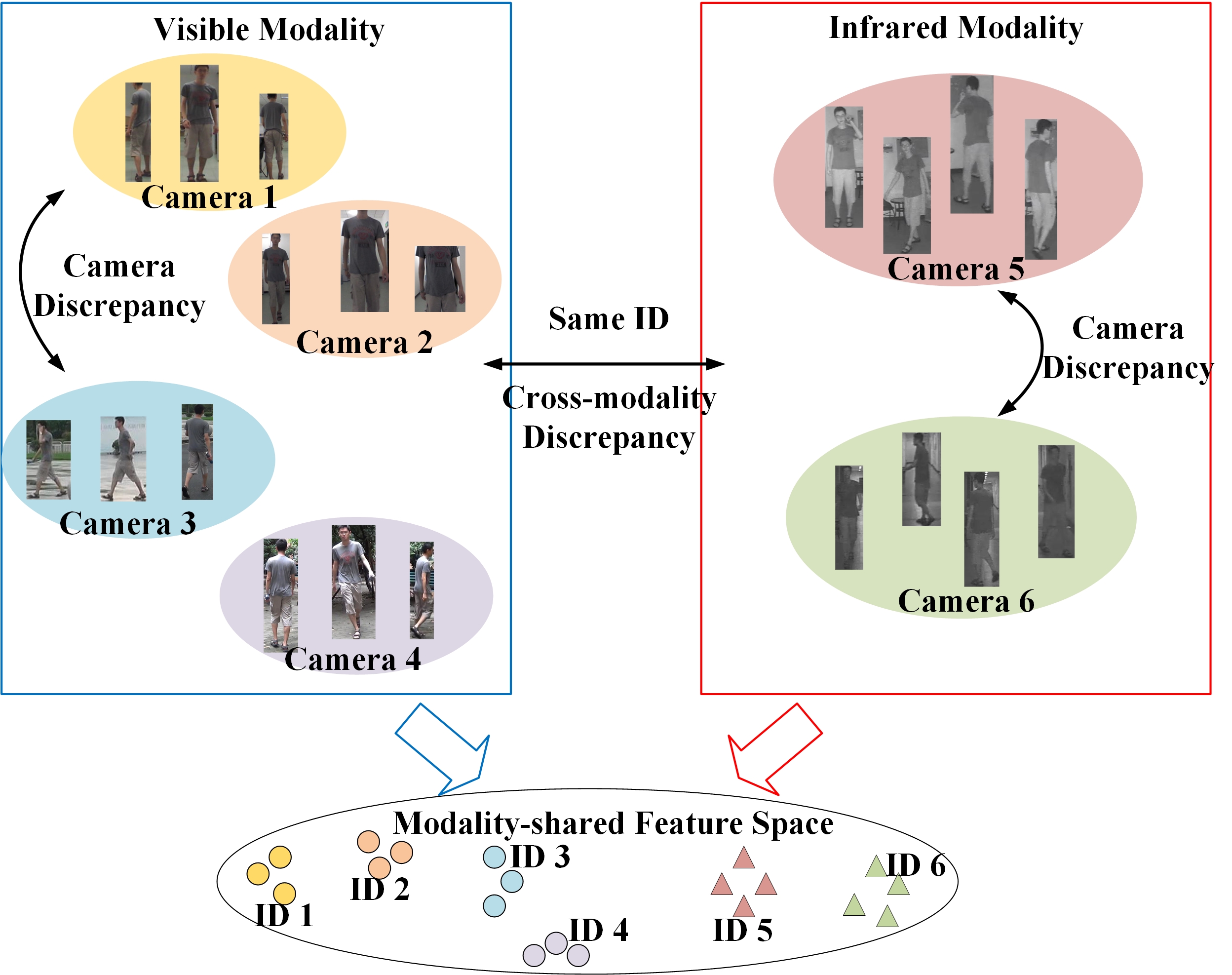}}
	\caption{Illustration of cross-modality and cross-camera discrepancies in clustering. Large variations caused by these discrepancies lead to identities splitting. Fine-tuning the network using these inaccurate labels obtains sub-optimal results.}
	\label{fig:camera_modal_clustering}       % Give a unique label
\end{figure}

The USL-VI-ReID method explores cross-modal associations, eliminating the requirement for manual identity labeling \cite{fan2018unsupervised,he2024dynamic}. Although DBSCAN \cite{DBSCAN_Ester_1996} is an effective clustering algorithm for pseudo-labeling unlabeled data, challenges still arise due to substantial cross-modality and cross-camera variations within the images from the same ground truth ID \cite{ye2020cross,zhong2020learning,zhong2018camstyle}. Fig. \ref{fig:camera_modal_clustering} illustrates the challenge in clustering cross-modal person data. Variations across cameras and modalities lead to excessive identity splitting and hinder accurate label assignment. Fine-tuning the network using these labels may amplify the distances within the same class more than those between different classes. In this case, additional noise can be potentially introduced to adversely affect the performance of the model. Existing methods \cite{yang2022augmented,wu2023unsupervised,pang2023cross,liang2021homogeneous,yang2023translation} primarily focus on eliminating differences between different modalities while neglecting the challenge of cross-camera discrepancy. For instance, previous approaches utilize graph matching \cite{wu2023unsupervised} and optimal transmission \cite{wang2022optimal} techniques to facilitate cross-modal cluster association. However, these methods are impeded by the severe issue of excessive identity splitting, which may affect the accuracy of the association.

\begin{figure*}
\centerline{\includegraphics[width=1.0\textwidth]{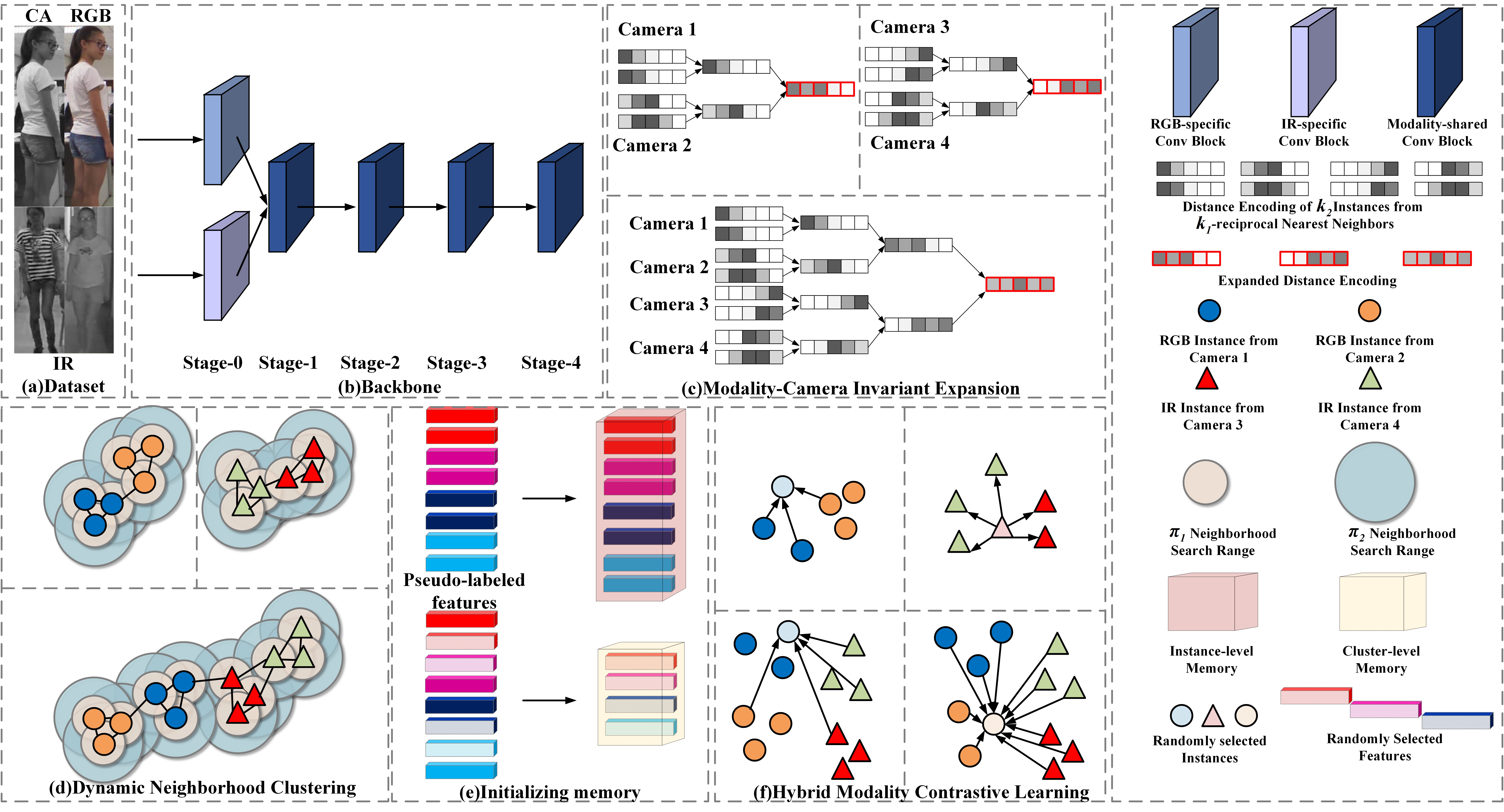}}
\caption{The flowchart of Dynamic Modality-Camera Invariant Clustering (DMIC) model. Our DMIC model is composed of three key modules: Modality-Camera Invariant Expansion (MIE), Dynamic Neighborhood Clustering (DNC), and Hybrid Modality Contrastive Learning (HMCL). MIE fuses the distance encoding from multiple cameras to generate modality-camera invariant embeddings for clustering. DNC employs two dynamic search strategies that optimize the network's performance, transitioning from improving discriminability to enhancing generalization. The estimated pseudo labels from MIE and DNC are used to initialize instance-level and cluster-level memories. HMCL includes intra-modality and inter-modality contrastive learning to learn modality-camera invariant representations. During the testing phase, our framework only utilizes the backbone for testing purposes.}
\label{fig:DMICflowchart}       % Give a unique label
\end{figure*}

To address the above problems, we propose a novel Dynamic Modality-Camera Invariant Clustering (DMIC) framework to eliminate cross-modality and cross-camera discrepancies at the clustering level. The flowchart of DMIC is displayed in Fig. \ref{fig:DMICflowchart}. To be specific, DMIC naturally combines Modality-Camera Invariant Expansion (MIE), Dynamic Neighborhood Clustering (DNC) and Hybrid Modality Contrastive Learning (HMCL) into a joint framework. The MIE integrates inter-modal and inter-camera distance coding, generating robust embeddings for the clustering algorithm. This eradicates the implicit inclusion of modality and camera information in the distance coding, resulting in modality-camera invariant embeddings. Consequently, modality-camera invariant associations can be established. To tackle the issue of intra-class distances surpassing inter-class distances due to excessive identity splitting, the DNC employs two dynamic search strategies. Specifically, for the first stragegy, we dynamically narrow the search radius to include the reliable positive samples within clusters, thereby enhancing the model's ability to distinguish between relevant and irrelevant samples. Subsequently, we dynamically broaden the search radius, using the model's refined discrimination to effectively incorporate reliable cross-camera and cross-modality samples into the cluster. The second strategy involves recalibrating the expanded distance coding. This refinement facilitates a broaden affinities of more cross-camera instances, which can be leveraged to contribute to cross-camera invariant learning. Importantly, our strategies do not require additional parameters as a cost, effectively enhancing the model's performance. Taking inspiration from \cite{yin2023real}, we design HMCL to optimize instance-level and cluster-level distributions. We randomly select instance from different modalities as cluster's centroid and update the representations of clusters in a real-time manner, which effectively reduces modal gap.

To sum up, the main contributions of this paper are list as follows:

\begin{itemize}
\item We propose a novel DMIC network for USL-VI-ReID that simultaneously eliminates both the cross-modality and cross-camera discrepancies in clustering.

\item The MIE fuses distance coding between inter-modal and inter-camera instances, which bridges cross-modality and cross-camera gaps.

\item The DNC consists of two dynamic search strategies that do not require additional parameters, facilitating early optimization of model discriminability and gradually extending generalization to different cameras and modalities.

\item The HMCL performs cluster-level and instance-level contrastive learning for intra-modality and inter-modality training. The cluster representations are updated using randomly selected cross-modal samples to obtain compact cross-modal distribution.

\item Extensive experiments on the SYSU-MM01 and RegDB datasets showcase the effectiveness and competitive performance of our DMIC framework.

\end{itemize}

\section{Related work}

In this section, we provide a brief overview of the following areas: Supervised Visible-Infrared person ReID (SVI-ReID), Unsupervised Single-Modality Person ReID (USL-ReID), and Unsupervised Learning Visible-Infrared Person Re-Identification (USL-VI-ReID).

\subsection{Supervised Visible-Infrared person ReID}
Supervised Visible-Infrared Person ReID (SVI-ReID) can be roughly divided into feature-level and image-level modality alignment methods.

The feature-level modality alignment methods aim to project cross-modal features into a shared subspace and employs metric learning techniques to narrow the distribution. In pursuit of this goal, Ye \textit{et al.} \cite{ye2020cross} introduce a modality-aware collaborative ensemble learning approach to eliminate modal discrepancies at both the instance and classifier levels. Lu \textit{et al.} \cite{lu2023tri} propose a two-step Wasserstein loss to align modality-unrelated information, which includes viewpoint, background, and posture. To fully leverage a wide array of cross-modality cues, a diverse embedding expansion network \cite{zhang2023diverse} is introduced to expand the feature set and reduce modal differences through triple-level constraints. Hao \textit{et al.} \cite{hao2021cross} design a camera-aware and modality-aware framework to enhance the discriminability and generalization of cross-modal representations.

Image-level modality alignment methods use generative networks or image enhancement techniques to eliminate modal differences between pixels. Ye \textit{et al.} \cite{ye2021channel} introduce a Channel exchangeable Augmentation (CA) method to enhance the generalization of visible stream. Li \textit{et al.} \cite{li2020infrared} design a lightweight network to transform visible images into $\mathit{X}$-modality in self-supervised manner. Zhang \textit{et al.} \cite{zhang2021towards} propose a non-linear simple generator to synthesize cross-modal images to middle modality. Furthermore, several works \cite{choi2020hi,wang2019learning,wang2019rgb,qi2023generative} successfully employ Generative Adversarial Networks (GANs) to achieve the transitions between visible and infrared modalities while preserving identity information. However, it is noted that modality alignment methods may inevitably introduce noise and may not be suitable for scenarios demanding high real-time performance.

\subsection{Unsupervised Single-Modality person ReID}
Unsupervised Single-Modality Person ReID (USL-ReID) tries to tackle the demanding and time-intensive task of assigning labels to visible modality images. Recently, mainstream methods fine-tune the network by assigning labels to the data through clustering algorithms. Memory-based learning methods are then used to optimize the relationships between instances and clusters. SPCL \cite{ge2020self} design unified contrastive learning to distinguish inter-cluster distributions. Cluster-Contrast \cite{dai2022cluster} refines InfoNCE \cite{he2020momentum} and presents ClusterNCE, effectively enhancing the optimization of unsupervised clusters and improving the performance of unsupervised methods. Lan \textit{et al.} \cite{lan2023learning} introduce multi-view features to contrastive framework, which enables efficient mining of partial cues and the refinement of pseudo labels. To solve camera discrepancy problem, Xuan \textit{et al.} \cite{xuan2022intra} divide unsupervised learning into inter-camera training and intra-camera training to generate reliable pseudo labels for cross-camera data. Zhang \textit{et al.} \cite{zhang2023camera} propose time-based camera contrastive learning to select the hardest camera centroid as a proxy for each cluster. ICE \cite{chen2021ice} design cross-camera proxy contrastive loss to mitigate camera discrepancy.

\subsection{Unsupervised Visible-Infrared person ReID}
The existing Unsupervised Learning Visible-Infrared Person Re-Identification (USL-VI-ReID) methods mainly focus on establishing cross-modal associations. ADCA \cite{yang2022augmented} adopt count priority selection method to facilitate cross-modal fusion. Wu \textit{et al.} \cite{wu2023unsupervised} design two cross-modal graphs to discover correspondences between different modalities. Pang \textit{et al.} \cite{pang2023cross} extract three channels from visible images, conduct clustering with infrared images, and subsequently utilize IoU for label refinement. Liang \textit{et al.} \cite{liang2021homogeneous} pretrain model with labeled single-modality dataset and introduce a homogeneous-to-heterogeneous training method. Wang \textit{et al.} \cite{wang2022optimal} utilize optimal transport techniques to transfer label knowledge from the visible modality to the infrared modality. However, the above-mentioned methods do not take into account the impact of camera differences in clustering phase. Especially, the susceptibility of visible images to camera discrepancy will lead to excessive identity splitting. In this case, using one-to-one or one-to-many association methods can exacerbate the impact of noisy labels. Unlike the above-mentioned methods, we fully utilize camera information throughout the clustering process to solve the identity splitting problem. Like ICE \cite{chen2021ice} and CAP \cite{wang2021camera}, GUR \cite{yang2023Toward} and DCCL \cite{yang2024Dual} cluster the samples within camera and employ cross-camera proxy contrastive loss to solve camera discrepancy. Unlike these methods, we do not need to perform clustering within individual cameras. Instead, we integrate camera information in a global clustering approach and introduce two dynamic search strategies to address camera differences.

\section{The Proposed Model}
	
In this section, we present a Dynamic Modality-Camera Invariant Clustering (DMIC) framework to simultaneously reduce cross-modality and cross-camera discrepancies. Our framework is illustrated in Fig. \ref{fig:DMICflowchart}.
\subsection{Problem Modeling}

In USL-VI-ReID, we adopt dual-stream backbone AGW \cite{ye2021deep} as backbone $f$. Given visible-infrared pedestrian datasets, we discard all identity labels, which means we are unable to use manual annotations as supervison. The visible-infrared pedestrian datasets can be represented as $\mathcal{D} =\{\mathcal{V} ,\mathcal{R} \}$, where $\mathcal{V}=\{x^v_i\}_{i=1}^{N_v}$ and $\mathcal{R}=\{x^r_i\}_{i=1}^{N_r}$ indicate visible and infrared images, respectively. $N_v$ denotes the number of visible images, and $N_r$ indicates the number of infrared images. We employ Channel exchangeable Augmentation (CA) \cite{ye2021channel} to enhance the generalization ability of visible stream. CA data can be represented as $\mathcal{C}=\{x^c_i\}_{i=1}^{N_v}$. Notably, we only enhance visible images using CA technique during the training process. 

It should be noted that within our framework, we include two training phases, i.e., intra-modality training and inter-modality training. The intra-modality training aims to enhance the initial discriminability of model, while the inter-modality training aims to develop the cross-modality and cross-camera generalization of model. For intra-modality training, we employ intra-modality clustering to assign modality-specific labels $\{\hat{y}^r,\hat{y}^v\}$ to the data in two different modalities, separately. In the context of inter-modality training, we utilize both intra-modality clustering and inter-modality clustering. Inter-modality clustering involves taking both infrared and visible data as inputs to the clustering algorithm, which assigns modality-shared labels $\{\hat{y}^m\}$. Therefore, during inter-modality training, each sample generates two pseudo-labels, i.e., $\{\hat{y}^r,\hat{y}^m\}$ for infrared samples and $\{\hat{y}^v,\hat{y}^m\}$ for visible samples.

\subsection{Modality-Camera Invariant Expansion}
In prior research \cite{yang2022augmented, wu2023unsupervised, liang2021homogeneous}, a commonly used clustering approach involved calculating Jaccard distances and utilizing distance encoding to generate embeddings for clustering infrared and visible data. Nevertheless, clusters within the visible modality tend to experience identity splitting due to significant variations in lighting and viewpoint caused by cross-camera discrepancy \cite{yang2023Toward}. It's worth noting that prior works mainly focus on enhancing the strategy of using the clustering algorithm, like  bottom-up clustering \cite{yang2023Toward,yang2024Dual,lin2019bottom}, with less comprehensive consideration of improving the algorithm itself. To address the problem of cross-camera and cross-modal variation, we introduce a simple and effective modal-camera invariant expansion (MIE) to improve the clustering algorithm in order to enhance the performance of clustering cross-camera and cross-modal samples.

Let's start with the method of obtaining distance encoding in existing clustering-based approaches. To obtain distance encoding, each feature is taken as a probe to compute the k-reciprocal encoding vector \cite{zhong2017re} with other features: 

\begin{equation}
	\begin{split}
		\mathcal{D}_i=[d_{i,1},d_{i,2},\cdots,d_{i,n}] 
	\end{split}
\end{equation}

\begin{equation}
	\begin{split}
		d_{i,j}=\begin{cases}
			exp({-\mathcal{M}(f_i,f_j)})& 
			\text{ if } f_j \in R(f_i,k_1) \\
			0& otherwise.
		\end{cases}
	\end{split}
\end{equation}
where $f_i$ and $f_j$ represent probe and gallery feature, respectively. The function $\mathcal{M}(\cdot,\cdot)$ represents the Mahalanobis distance, while $R(f_i, k_1)$ refers to the set of k-reciprocal nearest neighbors for $f_i$, with $k_1$ serving as a hyperparameter for adjusting the proximity range for distinguishing these nearest neighbors. 

The distance encoding of $f_i$ is then expanded by incorporating the distance encodings of the top-$k_2$ most similar gallery instances from its reciprocal neighbors. In this case, the distance encoding is effectively fused with contextual information from neighboring elements:
\begin{equation}
	\begin{split}
	\tilde{\mathcal{D}_i}=\frac{1}{k_2} {\textstyle \sum_{j=1}^{k_2}} \mathcal{D}_j 
		\label{eq:origin_fuse} 
	\end{split}
\end{equation}
where $k_2$ is smaller than $k_1$ to avoid introducing noisy instances, and $\tilde{\mathcal{D}_i}$ is the expanded distance encoding. Subsequently, we compute the Jaccard distance between the probe instance and other instances to obtain embedding for clustering:

\begin{equation}
	\begin{split}
		\mathcal{J}(f_i,f_j)=1-\frac{ {\textstyle \sum_{\pi =1}^{n}}min(d_{i,\pi},d_{j,\pi})}{\sum_{\pi =1}^{n}max(d_{i,\pi},d_{j,\pi})} 
	\end{split}
\end{equation}

\begin{equation}
	\begin{split}
		\mathcal{J}(f_i)=[\mathcal{J}(f_i,f_1),\mathcal{J}(f_i,f_2),\cdots,\mathcal{J}(f_i,f_n)]
	\end{split}
\end{equation}
where $min$ and $max$ operate the element-based minimization and maximization for two input vectors. $\mathcal{J}(f_i)$ denotes the embedding of $f_i$ for clustering. Then we take $\mathcal{J}=[\mathcal{J}(f_1),\mathcal{J}(f_2),\cdots,\mathcal{J}(f_n)]^\top$ as input to clustering algorithm \cite{DBSCAN_Ester_1996} for assigning pseudo labels $\hat{y}=\mathbf{DBSCAN}(\mathcal{J})$.

However, the majority of gallery instances are most similar to the query instances that have the same camera information. Consequently, the expanded distance encoding is unable to effectively capture the cross-camera neighborhood relationships. Thus, we equally fuse information from different cameras, ensuring that samples from different cameras can contribute equally to the expanded distance encoding. Eq. \ref{eq:origin_fuse} can be rewritten as:

\begin{equation}
	\begin{split}
	\tilde{\mathcal{D}_i}=\frac{1}{n^c} {\textstyle \sum_{j=1}^{n^c}} \mathcal{D}_j^{camera}
	\end{split}
\end{equation}

\begin{equation}
	\begin{split}
		\mathcal{D}_i^{camera}=\frac{1}{n^c_i} {\textstyle \sum_{j=1}^{k_2}} \textbf{1}\{l^{camera}_{j}=c\} \mathcal{D}_j
	\end{split}
\end{equation}
where $\textbf{1}\{\cdot\}$ is the indicator function, $n^c_i$ denotes the number of instance from camera $c$, $n^c$ indicates the number of camera, and $l^{camera}_j$ represents the camera label.

Notably, we employ MIE in the phase of intra-modality clustering and inter-modality clustering. Especially, we can bridge cross-modality discrepancy when fusing distance coding from visible camera domains and infrared camera domains. Unlike recent methods \cite{chen2021ice,yang2024Dual,yang2023Toward,Li2022camera,wang2021camera} focusing on camera differences, which employ camera proxies for contrastive learning, we introduce camera information during global clustering.

\subsection{Dynamic Neighborhood Clustering}
The clustering performance of USL-VI-ReID methods is critically influenced by hyperparameters in clustering algorithm, such as $eps$, $k_1$, and $k_2$. These hyperparameters affects clustering objectives, thereby influencing the optimization path of the network. Specifically, $eps$ defines the search radius for identifying neighborhoods, $k_1$ adjusts the proximity range to differentiate nearest neighbors, and $k_2$ determines the top-$k_2$ most similar gallery instances used for expanding distance encoding. However, in existing methods \cite{yang2022augmented,wu2023unsupervised,pang2023cross}, these hyperparameters are set empirically and remain constant during training. It is important to note that the choice of $eps$ determines the inclusion of noisy instances in the clusters. If $eps$ is set to be too large, it might result in including noisy instances, whereas if $eps$ is too small, it might lead to the exclusion of many valid cross-camera and cross-modality samples. Although some methods take into account the influence of $eps$, their consideration is not comprehensive. PUL \cite{fan2018unsupervised} maintains small $eps$ to select reliable samples, but this may not utilize valuable hard positive samples. And this method requires a predetermined number of identity categories, which is unknowable. DCCC \cite{he2024dynamic} only consider dynamic downsizing adjustment of eps and do not consider further scaling up to incorporate more hard positive samples with the help of network discriminatory power. Additionally, during the early stages of training, the model's discriminative capability is limited, and setting a higher value for $k_2$ may incorporate inaccurate correlation data into the expanded distance coding. On the contrary, maintaining an appropriate value of $k_2$ during the middle and late stages of training prevents the aggregation of distance coding with cross-modality and cross-camera samples that exhibit relatively low similarity. The dynamic adjustment of $k_2$ is not considered in current methods.

In this subsection, we introduce Dynamic Neighborhood Clustering (DNC) to dynamically adjust the clustering objective. We anticipate that DNC, in collaboration with MIE, can effectively address the identity splitting issue. First and foremost, we begin with an assumption: during the early stages of model optimization (intra-modality training), we aim to minimize the inclusion of noisy instances within clusters to enhance the model's discriminative power. As the optimization progresses into the middle and later stages (inter-modality training), our objective shifts towards gradually incorporating cross-modality and cross-camera positive samples into the clusters to improve the model's ability to generalize. To achieve this, we implement dynamic exponential schedulers for $eps$ and $k_2$:

\begin{equation}
	\begin{split}
		\pi_1=\pi_2*\sigma ^{epochs}_n
	\end{split}
	\label{eq:eps_smaller}
\end{equation}

\begin{equation}
	\begin{split}
		\pi_2=\pi_1*\sigma ^{epochs}_b
	\end{split}
	\label{eq:eps_larger}
\end{equation}

\begin{equation}
	\begin{split}
		\epsilon _2=\epsilon _1*\sigma ^{epochs}_k
	\end{split}
	\label{eq:k2_larger}
\end{equation}
where $\sigma_n\in[0,1)$ denote the decay ratio, while $\sigma_b\in(1,+\infty]$ and $\sigma_k \in (1,+\infty] $ indicate growth ratio. $\pi_2$ is the upper bound and $\pi_1$ is the lower bound of $eps$. $\epsilon _2$ $(>\epsilon _1)$ represents the upper bound of $k_2$. To better illustrate the use of dynamic strategies for intra-modality and inter-modality training, we initially define the estimation of modality-specific labels $\{\hat{y}^r,\hat{y}^v\}$:

\begin{equation}
	\begin{split}
		\hat{y}^r=\mathbf{DBSCAN}(\mathcal{J}^r;k_1,k_2,eps)\\
		\hat{y}^v=\mathbf{DBSCAN}(\mathcal{J}^v;k_1,k_2,eps)
	\end{split}
	\label{eq:intraclustering} 
\end{equation}
where $\mathcal{J}^r$ and $\mathcal{J}^v$ are embeddings obtained from MIE. Please note that $k_1$ and $k_2$ are not hyperparameters for DBSCAN. They are hyperparameters calculated within MIE. However, we include them here for clarity in our further demonstration. Then, we also define estimation of modality-share labels $\{\hat{y}^m\}$:
\begin{equation}
	\begin{split}
		\hat{y}^m=\mathbf{DBSCAN}(\mathcal{J}^m;k_1,k_2,eps)
	\end{split}
	\label{eq:interclustering} 
\end{equation}
where $\mathcal{J}^m$ is the embedding of infrared and visible data together as input for the MIE module.

\begin{figure}
	\centerline{\includegraphics[width=0.5\textwidth]{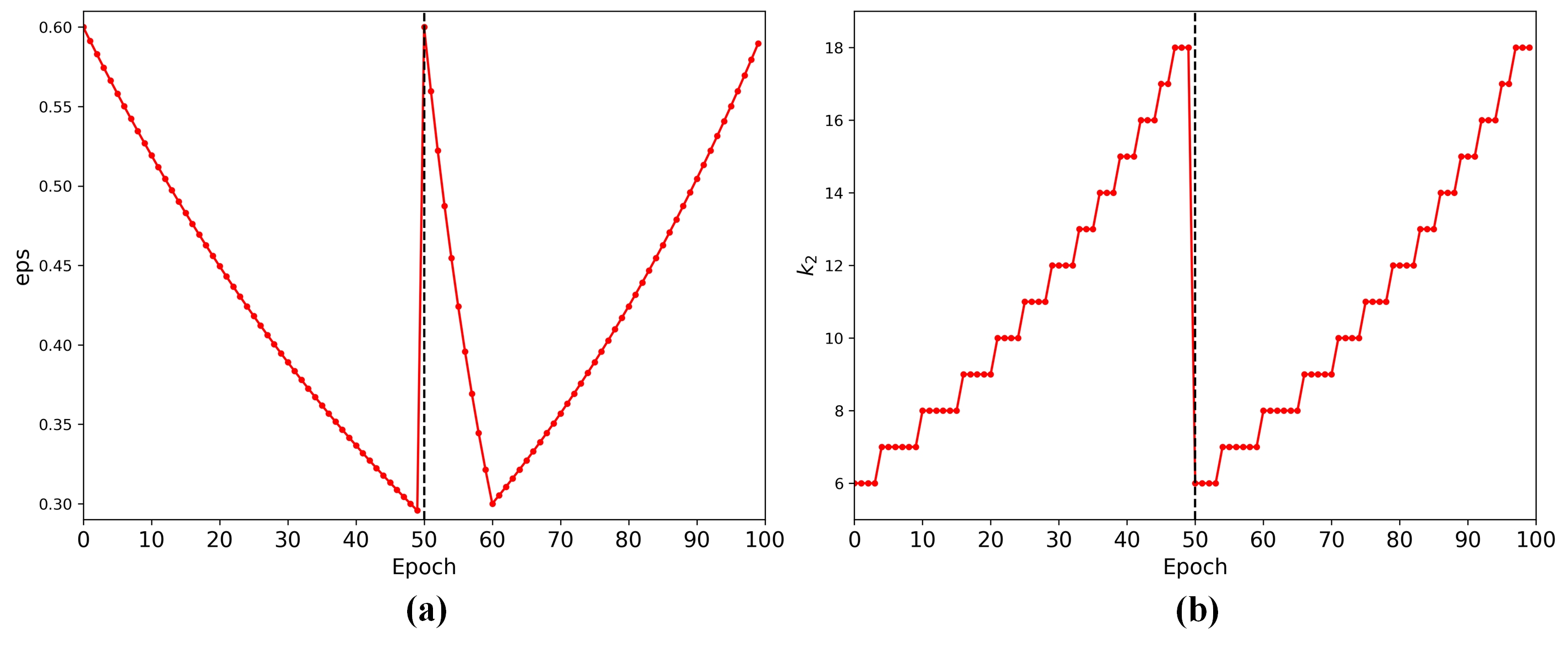}}
	\caption{Dynamic schedular in DNC. (a) is the dynamic schedular for $eps$ in Eq. \ref{eq:intraclustering} and Eq. \ref{eq:interclustering}, while (b) is for $k_2$ in Eq. \ref{eq:intraclustering}.}
	\label{fig:dynamic_strategy}       % Give a unique label
\end{figure}

\begin{figure}
	\centerline{\includegraphics[width=0.5\textwidth]{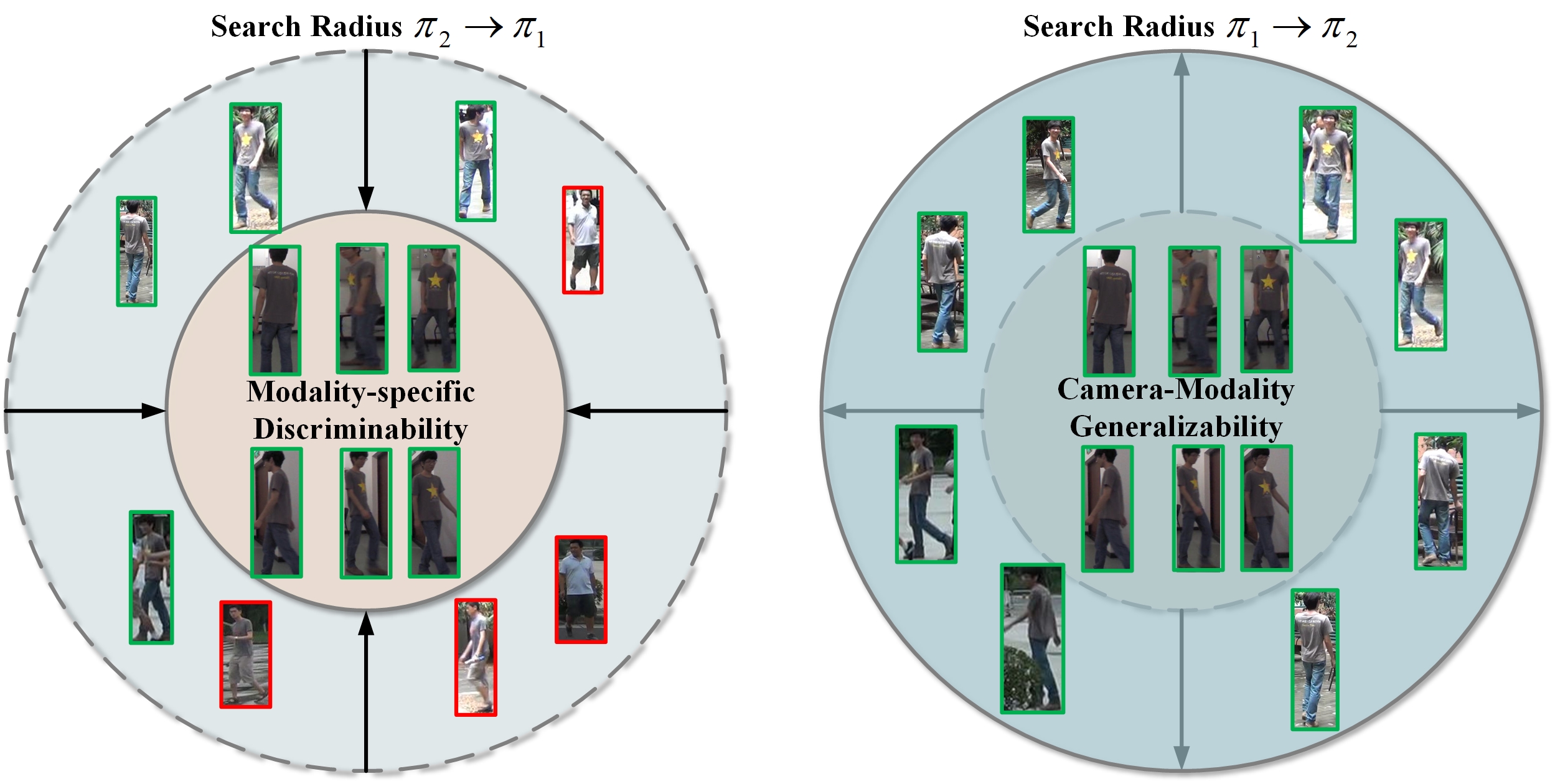}}
	\caption{Illustratiion of the clustering results of DNC. Taking $eps$ for an example, we establish upper and lower dynamic range limits for $eps$ denoted as $\pi_2$ and $\pi_1$, respectively. Initially, $eps$ decreases from $\pi_2$ to $\pi_1$, excluding noisy instances from clusters. Subsequently, $eps$ expands from $\pi_1$ to $\pi_2$, progressively incorporating cross-modality and cross-camera instances into clusters. In this manner, the model first improves discriminability and then gradually develops cross-modality and cross-camera generalizability. }
	\label{fig:DNC}       % Give a unique label
\end{figure}

As previously mentioned, intra-modality training focuses on improving the initial discrimination capabilities of the model, whereas inter-modality training aims to enhance the model's ability to generalize across different modalities and cameras. Consequently, we employ distinct dynamic strategies for these two stages of training to accomplish this objective, as shown in Fig. \ref{fig:dynamic_strategy}. \textbf{(1) Intra-modality training:} The $eps$ in Eq. \ref{eq:intraclustering} undergoes large-to-small process according to Eq. \ref{eq:eps_smaller}, while $k_2$ in Eq. \ref{eq:intraclustering} experiences the small-to-large process according to Eq. \ref{eq:k2_larger}. With regards to $eps$, the clustering algorithm systematically eliminates noisy samples. For $k_2$, the clustering algorithm gradually explores affinities within intra-camera instances and subsequently transitions to exploring affinities between inter-camera instances in collaboration with $eps$. \textbf{(2) Inter-modality training:} The $eps$ in Eq. \ref{eq:intraclustering} and Eq. \ref{eq:interclustering} first go through a small period of epochs (set to 10 epochs) from large to small, and then go through the rest (set to 40 epochs) from small to large. The small-to-large stage can serve as a warm-up process for cross-modal clustering, aiding in the separation of some solid modality-specific clusters. In the small-to-large phase for $eps$, the clustering algorithm progressively incorporates reliable cross-modality and cross-camera instances into clusters. However, the adjustment of $k_2$ in Eq. \ref{eq:intraclustering} and Eq. \ref{eq:interclustering} follows distinct paths. The former experiences the small-to-large process, while the latter remains the a larger value of $\epsilon _3$. Combining visible and infrared samples in the clustering algorithm requires a relatively large value for $k_2$ to enhance the likelihood of linking cross-modal samples with substantial modal differences. This is essential for maintaining effective connections between visible and infrared samples in the clustering algorithm, achieved by keeping a relatively large value for $k_2$.

Overall, the dynamic adjustment of $eps$ serves to illustrate the purpose of DNC, which can be visualized in Fig. \ref{fig:DNC}. The clustering method plays a pivotal role in unsupervised approaches, directly influencing network performance. Our DNC dynamically fine-tunes the clustering objective to control the network's learning objectives. Notably, DNC cooperates with MIE, effectively addressing the challenge of identity splitting resulting from variations in camera and modal distributions. Further details regarding the effectiveness of DNC are presented in the experiment section.

\subsection{Hybrid Modality Contrastive Learning}
With the intra-modal pseudo labels $\{\hat{y}^r,\hat{y}^v\}$ and inter-modal pseudo labels $\{\hat{y}^m\}$ obtained from MIE and DNC, we propose Hybrid Modality Contrastive Learning (HMCL) to refine the distributions between clusters and instances.

Unsupervised methods commonly take refined InfoNCE \cite{dai2022cluster,he2020momentum} as their loss function, which can be defined as:
\begin{equation}
	\begin{split}
		\mathcal{L} =-\sum_{i=1}^{P\times Z} log \frac{exp(q_i\cdot \phi[\hat{y}_i]/\tau )}{ {\textstyle \sum_{k=0}^{I}}exp(q_i\cdot  \phi[k]/\tau) } 
	\end{split}
\end{equation}
where $P$ and $Z$ indicate the number of sampled individuals and instances per individual, respectively. $q_i$ denotes L2-normalized query instance in training mini-batch. $\phi[\hat{y}_i]$ is the positive cluster representation of $q_i$ and $\phi[k]$ represents each cluster representation stored in memory, where cluster representations are obtained by averaging the instance features in the clusters. $\tau$ is the temperature factor and $I$ is the total number of clusters. Then a momentum updating strategy \cite{he2020momentum} is employed to update the memory after each iteration:

\begin{equation}
	\begin{split}
		\phi[\hat{y}_i] \gets \lambda \phi[\hat{y}_i] + (1-\lambda) q_i \quad (i=1,2,...,P\times Z)
	\end{split}
\end{equation}

Similar to \cite{yin2023real}, we adopt a strategy of randomly sampling instances to update cluster representations. Subsequently, we introduce cluster-level and instance-level losses to refine global and partial distributions, respectively.

For intra-modality training, we construct two kinds of memories, i.e., cluster-level and instance-level memories:

\begin{equation}
	\begin{split}
		\phi_I^r[i] &= f_i^r\\
		\phi_I^v[i] &= f_i^{v}\\
		\phi_C^r[\hat{y}_i^r] &= \frac{1}{\left | O^r_{\hat{y}^r_i}  \right | } \sum_{f^r_i\in O^r_{\hat{y}^r_i}}^{} f_i^r\\
		\phi_C^v[\hat{y}_i^v] &=\frac{1}{\left | O^v_{\hat{y}^v_i}  \right | } \sum_{f^v_i\in O^v_{\hat{y}^v_i}}^{} f_i^v
	\end{split}
	\label{eq:memory} 
\end{equation}
where $\phi_I^r$ and $\phi_I^v$ represent instance-level memories for infrared and visible modalities, while $\phi_C^r$ and $\phi_C^v$ indicate cluster-level memories. $ O^{r(v)}_{\hat{y}^{r(v)}_i}$ denotes the $\hat{y}^{r(v)}_i$-th cluster set in infrared or visible modality, and $\left |\cdot \right |$ represents the number of instances in specific cluster. To establish a stable starting point for optimization, we initialize the cluster-level memory by taking the average value. After each iteration, we randomly select instances to update memory instead of employing momentum updating strategy, ensuring the real-time update of memory. In the case of the visible modality, to fully leverage the CA modality, we randomly update visible memory using CA and visible features, specifically $\phi_I^v[i] \gets f_i^{(v,c)}$ and $\phi_C^v[\hat{y}_i^v] \gets f_i^{(v,c)}$. As for infrared modality, we update memory by  $\phi_I^r[i] \gets f_i^r$ and $\phi_C^r[\hat{y}_i^r] \gets f_i^r$.

The cluster-level contrastive loss for visible and infrared modalities can be expressed as:

\begin{align}
		\mathcal{L}^v_C &=-\sum_{i=1}^{2\times P\times Z} log \frac{exp(q_i^{(v,c)}\cdot \phi_C^v[\hat{y}_i^v]/\tau )}{ {\textstyle \sum_{k=0}^{I}}exp(q_i^v\cdot   \phi_C^v[k]/\tau) } \\
		\mathcal{L}^r_C &=-\sum_{i=1}^{P\times Z} log \frac{exp(q_i^r\cdot \phi_C^r[\hat{y}_i^r]/\tau )}{ {\textstyle \sum_{k=0}^{I}}exp(q_i^r\cdot  \phi_C^r[k]/\tau) } 
\end{align}
where $q_i^{(v,c)}$ denote the query features in the mini-batch from visible and CA modalites, and $q_i^r$ indicates the query feature from infrared modality.

To explore the instance relationships during the training process, the instance-level contrastive loss can be formulated as:
\begin{align}
		\mathcal{L}^v_I &=-\sum_{i=1}^{2 \times P\times Z} log \frac{exp(q_i^{(v,c)}\cdot \phi_I^v[\hat{y}_i^v]/\tau )}{ {\textstyle \sum_{k=0}^{I}}exp(q_i^v\cdot   \phi_I^v[k]/\tau) } \\
		\mathcal{L}^r_I  &=-\sum_{i=1}^{P\times Z} log \frac{exp(q_i^r\cdot \phi_I^r[\hat{y}_i^r]/\tau )}{ {\textstyle \sum_{k=0}^{I}}exp(q_i^r\cdot  \phi_I^r[k]/\tau) } 
\end{align}

For inter-modality training, we enhance the training process by introducing global cluster-level memory and instance-level memory, in addition to the intra-modality training, denoted as:
\begin{equation}
	\begin{split}
		\phi_I^m[i] &=  f_i^{(v,r)}\\
		\phi_C^m[\hat{y}_i^m] &= \frac{1}{\left | O^m_{\hat{y}^m_i}  \right | } \sum_{f^{(v,r)}_i\in O^m_{\hat{y}^m_i}}^{} f_i^{(v,r)}
	\end{split}
\end{equation}
where $ O^m_{\hat{y}^m_i}$ denotes the $\hat{y}^{m}_i$-th cluster set according to DBSCAN in the process of inter-modality clustering. When updating these memories, CA modality features can be included through random selection, as indicated by $\phi_I^m[i] \gets f_i^{(v,r,c)}$ and $\phi_C^m[i] \gets f_i^{(v,r,c)}$.

The cluster-level and instance-level contrastive losses for inter-modality training can be expressed as:

\begin{align}
	\mathcal{L}^m_C &= -\sum_{i=1}^{3 \times P\times Z} \log \frac{\exp(q_i^{(v,r,c)}\cdot \phi_C^m[\hat{y}_i^m]/\tau )}{\sum_{k=0}^{I}\exp(q_i^{(v,r,c)}\cdot \phi_C^m[k]/\tau)} \\
	\mathcal{L}^m_I &= -\sum_{i=1}^{3 \times P\times Z} \log \frac{\exp(q_i^{(v,r,c)}\cdot \phi_I^m[\hat{y}_i^m]/\tau )}{\sum_{k=0}^{I}\exp(q_i^{(v,r,c)}\cdot \phi_I^m[k]/\tau)}
\end{align}
where $q_i^{(v,r,c)}$ indicate that the query features are sampled from visible, infrared and CA modalities.

During the training process, HMCL firstly perform intra-modality learning and explore reliable intra-modal clusters. This preference stems from the simplicity of initiating intra-modal learning before inter-modal learning. The implementation of direct inter-modal clustering results in an excessive introduction of noise. Such an increase in noise, especially during the early stages of model training, accumulates progressively, thereby negatively impacting the final efficacy of the model. As a result, transitioning from intra-modal to inter-modal learning aids in steering the model toward a more organized optimization path. Moreover, with the combined influence of MIE and DNC, cross-modal discrepancies are eliminated. Simultaneously, cross-camera information steadily aggregates, providing robustness against cross-camera disturbances.
\begin{table}
\begin{spacing}{1.0}
\footnotesize
\centering{\begin{tabular}{p{8.5cm}l}
  \hline
\textbf{Algorithm 1:} \quad DMIC approach\\\hline

\textbf{Input:} \ Unlabeled infrared training data $\mathcal{R}$, unlabeled visible training data $\mathcal{V}$, epochs number $epoch$ and training iterations number $iter$.\\

\textbf{Output:} \ Backbone network parameters $\theta$.\\

1: \textbf{for} $\kappa=1,2,...,epoch$ \textbf{do} $\textcolor[rgb]{0,0,1}{\mathrm{\#Intra-modality\quad training}}$\\

2: \quad Adjust $eps$ and $k_2$ by Eq. (8)-(10); 

3: \quad Generate intra-modality pseudo labels $\{\hat{y}^r,\hat{y}^v\}$ by Eq. (1), (2), (4), (5), (6), (7) and (11);\\

4: \quad Initialize memories $\{\phi^{r}_I,\phi^{v}_I,\phi^{r}_C,\phi^{v}_C\}$ by Eq.(15);\\

5: \quad  \textbf{for} $\pi  =1,2,...,iter$ \textbf{do}\\

6: \quad \quad Update $\theta$ by minimizing Eq. (16)-(19); \\

7: \quad \quad Update memories $\{\phi^{r}_I,\phi^{v}_I,\phi^{r}_C,\phi^{v}_C\}$.\\

8: \quad \textbf{end for}\\

9: \textbf{end for}\\

10: \textbf{for} $\kappa=1,2,...,epoch$ \textbf{do} $\textcolor[rgb]{0,0,1}{\mathrm{\#Inter-modality\quad training}}$\\

11: \quad Adjust $eps$ and $k_2$ by Eq. (8)-(10);

12: \quad Generate intra-modality labels $\{\hat{y}^r,\hat{y}^v\}$ and inter-modality pseudo labels $\{\hat{y}^m\}$ by Eq. (1), (2), (4), (5), (6), (7), (11) and (12);\\

13: \quad Initialize memories $\{\phi^{r}_I,\phi^{v}_I,\phi^{m}_I,\phi^{r}_C,\phi^{v}_C,\phi^{m}_C\}$ by Eq. (15) and (20);\\

14: \quad  \textbf{for} $\pi  =1,2,...,iter$ \textbf{do}\\

15: \quad \quad Update $\theta$ by minimizing Eq. (16)-(19) and (21)-(22); \\

16: \quad \quad Update memories $\{\phi^{r}_I,\phi^{v}_I,\phi^{m}_I,\phi^{r}_C,\phi^{v}_C,\phi^{m}_C\}$.\\

17: \quad \textbf{end for}\\

18: \textbf{end for}\\

19: \textbf{Return} $\theta$\\
\hline
\end{tabular}}
\end{spacing}
\end{table}

\begin{figure}
\centerline{\includegraphics[width=0.4\textwidth]{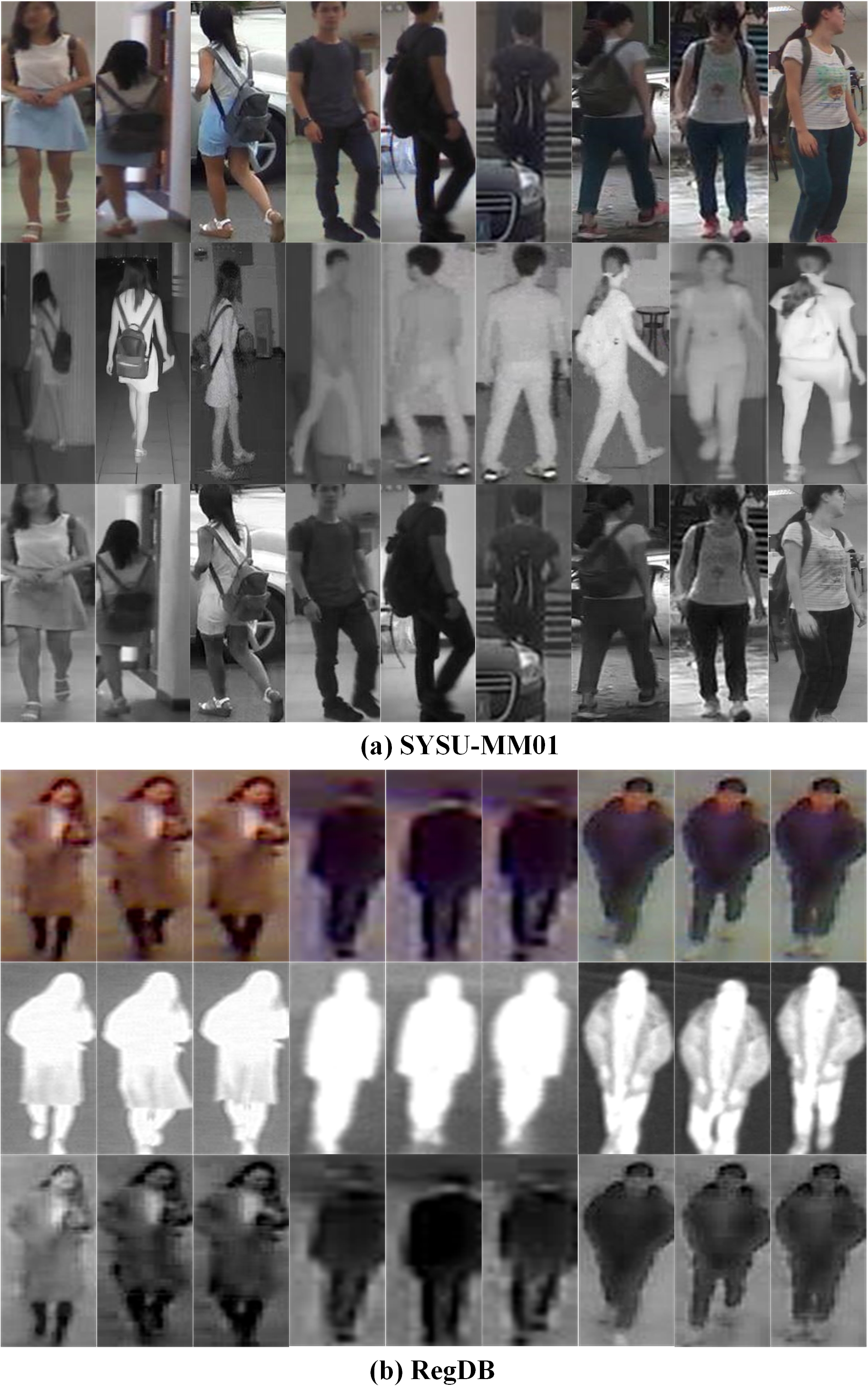}}
\caption{Example images from visible-infrared pedestrian databases. The images from the upper row are in visible modality, the images from the second row are in infrared modality, and the images from the last row are in CA modality.}
\label{fig:crop_two_datasets}       % Give a unique label
\end{figure}

\subsection{Optimization Algorithm}
The proposed DMIC network naturally integrates MIE, DNC and HMCL into a unified framework. These key modules can establish a synergistic effect to bolster the model's robustness against cross-modality and cross-camera discrepancies. Since we have both intra-modal and inter-modal training stages, the overall loss function for DMIC can be formally expressed as:

\begin{equation}
\begin{split}
\mathcal{L} =\varphi_1 (\mathcal{L}^r_I + \mathcal{L}^v_I+\mathcal{L}^m_I) + \varphi_2 (\mathcal{L}^r_C + \mathcal{L}^v_C+\mathcal{L}^m_C)
\end{split}
\end{equation}

\noindent where $\varphi_1$ and $\varphi_2$ denote the trade-off weights for the cluster-level and instance-level loss functions. The optimization process of our method is outlined in Algorithm 1, with $epoch$ representing the number of epochs, $\kappa$ signifying the epoch index, and $iter$ indicating the number of training iterations.

\section{Experiments Evaluation}

In this section, we first introduce two datasets, i.e., SYSU-MM01 \cite{wu2017rgb} and RegDB \cite{nguyen2017person}, and experiment implementation. Then, we conduct experiments to compare our method with state-of-the-art methods on these two public datasets, which shows our competitive performance. Finally, we conduct several property analyses to demonstrate the impact of each component.

\subsection{Datasets and Settings}

\textbf{SYSU-MM01.}\quad SYSU-MM01 \cite{wu2017rgb} is a widely used dataset for evaluating recently published methods. This dataset consists of 287,628 visible and 15,792 infrared images, which are captured from 491 subjects. The training set includes 395 identities, while 96 identities are adopted for testing. Importantly, there is no overlap between the training and testing sets. The collected images are from six kinds of cameras, which include 4 RGB cameras and 2 infrared cameras. Therefore, as shown in the Fig. \ref{fig:crop_two_datasets} (a), there are significant variations in pedestrians' pose, viewpoint, and background. These pose considerable challenges for unsupervised clustering algorithms. Following \cite{ye2021deep,yang2022augmented}, the dataset can be divided into two different settings, i.e., all search and indoor search.

\textbf{RegDB.}\quad RegDB \cite{nguyen2017person} is a relatively small-scale visible-infrared dataset. This dataset includes 8,240 images from 412 identities. Following \cite{ye2021deep,yang2022augmented}, a non-overlapping selection of 206 identities is employed for both training and testing purposes. Each identity consists of an equal distribution of 10 visible and 10 thermal images. As displayed in Fig. \ref{fig:crop_two_datasets} (b), images are captured from consecutive frames, resulting in reduced variation among images of the same identity. This dataset contains ten protocols, and the final performance is taken from the overall average accuracy.

\textbf{Metrics.}\quad Cumulated Matching Characteristics (CMC), Mean Average Precision (mAP) and Mean Inverse Negative Penalty (MINP) \cite{ye2021deep} are employed to fairly compare the performance of our method with existing methods.

\begin{table*}[t]
	\centering
	\renewcommand{\multirowsetup}{\centering}
	\renewcommand{\arraystretch}{1.3}
		\caption{The performance(\%) comparison on SYSU-MM01 dataset. \dag indicates training with 128 batch size. }\label{tab:comparison_sysu}
	\scalebox{0.8}
	{\setlength{\tabcolsep}{2.0 mm}{\begin{tabular}{c|cc|ccccc|ccccc}
				\hline
				& \multicolumn{2}{c|}{SYSU-MM01 Settings} & \multicolumn{5}{c|}{All Search}                 & \multicolumn{5}{c}{Indoor Search}               \\ [-1ex]
				& Methods                & Venue          & Rank-1 & Rank-10 & Rank-20 & mAP & mINP & Rank-1 & Rank-10 & Rank-20 & mAP & mINP \\ [-1ex] \hline\hline 
				\multirow{9}{*}{\rotatebox{90}{Supervised}}  
				& Zero-Padding \cite{wu2017rgb}             & ICCV-17        & 14.80   & 54.12    & 71.33   &15.95    & -        & 20.58      & 68.38       & 85.79       & 26.92       & -        \\ [-0.5ex]  
				& eBDTR \cite{ye2019bi}             & TIFS-19        & 27.82   & 67.34    & 81.34   &58.42    & -        & 32.46      & 77.42       & 89.62       & 42.46      & -        \\ [-0.5ex]  
				& X-Modal \cite{li2020infrared}             & AAAI-20        & 49.9   & 89.8    & 96.0    & 50.7    & -        & -      & -       & -       & -       & -        \\ [-0.5ex]
				& Hi-CMD \cite{choi2020hi}                 & CVPR-20        & 34.9   & 77.6    & -       & 35.9    & -        & -      & -       & -       & -       & -        \\ [-0.5ex]

				& AGW \cite{ye2021deep}                   & TPAMI-21       & 47.50  & 84.39   & 92.14   & 47.65   & 35.30    & 54.17  & 91.14   & 95.98   & 62.97   & 59.23    \\[-0.5ex]
				& DDAG \cite{choi2020hi}                 & ECCV-20        & 54.75   & 90.39    & 95.81       & 53.02    &39.62        &61.02      & 94.06      & 98.41       & 67.98       & 62.61        \\ [-0.5ex]
				& VCD+VML \cite{tian2021farewell}                & CVPR-21        & 60.02  & 94.18   & 98.14   & 58.80   & -        & 66.05  & 96.59   & 99.38   & 72.98   & -        \\[-0.5ex]
				& CA \cite{ye2021channel}                    & ICCV-21        & 69.88  & 95.71   & 98.46   & 66.89   & 53.61    & 76.26  & 97.88   & 99.49   & 80.37   & 76.79    \\[-0.5ex]
				& MPANet \cite{ye2021channel}                    & CVPR-21        & 70.58  & 96.21   & 98.80  & 68.24   & -    & 76.74  &98.21 & 99.57   & 80.95   & -   \\[-0.5ex]
				& MSO \cite{gao2021mso}                   & MM-21          & 58.70  & 92.06   & -       & 56.42   & -        & 63.09  & 96.61   & -       & 70.31   & -        \\[-0.5ex]
				& MCLNet \cite{hao2021cross}                & ICCV-21        & 65.40  & 93.33   & 97.14   & 61.98   & 47.39    & 72.56  & 96.98   & 99.20   & 76.58   & 72.10    \\[-0.5ex]
				& SPOT \cite{chen2022structure}                & TIP-22        & 65.34  & 92.73  & 97.04   & 62.25   & 48.86    & 69.42  & 96.22   & 99.12   & 74.63   & 70.48    \\[-0.5ex]
				& DMiR \cite{hu2022adversarial}                & TCSVT-22        & 50.54  & 88.12  & 94.84   & 49.29   & -    & 53.92  & 92.50  & 97.09  & 62.49   & -    \\[-0.5ex]
								
				& FMCNet \cite{zhang2022fmcnet}                & CVPR-22        & 66.34  & -       & -       & 62.51   & -        & 68.15  & -       & -       & 74.09   & -        \\[-0.5ex]
				& MAUM \cite{liu2022learning}                  & CVPR-22        & 71.68  & -       & -       & 68.79   & -        & 76.97  & -       & -       & 81.94   & -        \\ [-0.5ex]
				& TMD \cite{lu2023tri}                  & TMM-23        & 68.81  & 93.08       & 96.84       &63.96   & 48.11        & 76.31  & 97.28       & 98.91       & 74.52   & 65.05        \\ [-0.5ex]\hline\hline

				\multirow{16}{*}{\rotatebox{90}{Unsupervised}}    
				& SSG  \cite{fu2019self}                 & ICCV-19        & 2.32  & 17.23   & 28.88   & 5.00   & -   &- & -   & -   & -   & -    \\
				& ECN  \cite{zhong2019invariance}                 & CVPR-19        & 8.07  &32.49   & 45.95   & 12.68  & -    & -  & -   & -   & -   & -   \\
				& SPCL  \cite{ge2020self}                 & NIPS-20        & 18.37  & 54.08   & 69.02   & 19.39   & 10.99    & 26.83  & 68.31   & 83.24   & 36.42   & 33.05    \\
				& MMT  \cite{ge2020mutual}                  & ICLR-20        & 21.47  & 59.65   & 73.29   & 21.53   & 11.50    & 22.79  & 63.18   & 79.04   & 31.50   & 27.66    \\[-0.5ex]
				& CAP \cite{wang2021camera}                   & AAAI-21        & 16.82  & 47.6    & 61.42   & 15.71   & 7.02     & 24.57  & 57.93   & 72.74   & 30.74   & 26.15    \\[-0.5ex]
				& IICS  \cite{xuan2021intra}                 & CVPR-21        & 14.39  & 47.91   & 62.32   & 15.74   & 8.41   & 15.91  & 54.20   & 71.49   & 24.87   & 22.15   \\		
				& ICE  \cite{chen2021ice}                 & ICCV-21        & 20.54  & 57.50  & 70.89   & 20.39   & 10.24    & 29.81  & 69.41   & 82.66   & 38.35  & 34.32    \\

				& Cluster-Contrast \cite{dai2022cluster}      & ACCV-22           & 20.16  & 59.27   & 72.5    & 22.00   & 12.97    & 23.33  & 68.13   & 82.66   & 34.01   & 30.88    \\[-0.5ex]
				& PPLR   \cite{cho2022part}                & CVPR-22        & 12.58  & 47.43   & 62.69   & 12.78   & 4.85     & 13.65  & 52.66   & 70.28   & 22.19   & 18.35    \\[-0.5ex]
				& ISE \cite{zhang2022implicit}                   & CVPR-22        & 20.01  & 57.45   & 72.50   & 18.93   & 8.54     & 14.22  & 58.33   & 75.32   & 24.62   & 21.74    \\  & H2H  \cite{liang2021homogeneous}                  & TIP-21         & 30.15  & 65.92   & 77.32   & 29.40   & -        & -      & -       & -       & -       & -        \\[-0.5ex]
				& OTLA \cite{wang2022optimal}                  & ECCV-22        & 29.98  & 71.79   & 83.85   & 27.13   & -        & 29.8   & -       & -       & 38.8    & -        \\[-0.5ex]
				& ADCA  \cite{yang2022augmented}                 & MM-22          & 45.51  & 85.29   & 93.16   & 42.73   & 28.29    & 50.60  & 89.66   & 96.15   & 59.11   & 55.17    \\[-0.5ex]  

				& CHCR \cite{pang2023cross}   & TCSVT-23   & 47.72      &  87.29       &   94.23     &   45.34     &   -      &   50.12     &  90.78       & 95.60      &  42.17      &  -        \\ [-0.5ex] 
				& TAA \cite{yang2023translation}   & TIP-23   & 48.77     & 85.69       &   93.32    &   42.43    &   25.37      &   50.12     &  87.11       & 94.55      &  56.02      &  49.96        \\ [-0.5ex] 
				& PGMAL \cite{wu2023unsupervised}   & CVPR-23   & 57.27       &  92.48       &   97.23     &   51.78      &   34.96      &   56.23     &  90.19       &  95.39       &   62.72      &  58.13   
				     \\ [-0.5ex] 
   				& CCLNet \cite{chen2023Unveiling}   & MM-23   &54.03      &88.80      &   95.02     &   50.19     &   -      &   56.68   & 91.14     &  97.23      &  65.12     &  -       \\ [-0.5ex]
				& GUR \cite{yang2023Toward}   & ICCV-23   &63.51     & -      &   -      &  61.63      &   47.93     &   71.11   & -      &  -       &  76.23    &  72.57     \\ [-0.5ex]
				& DCCL \cite{yang2024Dual}   & TIFS-24   &63.18      & -      &   -      &  58.62   &  42.99   &   66.67   & -      &  -       &  71.82      & 67.46     \\ [-0.5ex] \hline\hline
				& $\mathbf{DMIC (ours)}$     & -       & $60.26$       &   $92.67$     &    96.95      &   $57.82$      &   $44.13$       & $65.83$        &  $96.70$       &      $99.11$   &    $72.55$    &  $68.60$     \\ [-0.5ex]
				& $\mathbf{DMIC^\dag (ours)}$     & -       & $\mathbf{ 65.24}$       &   $\mathbf{93.95}$     &   $\mathbf{97.69}$      &   $\mathbf{62.01}$      &   $\mathbf{48.00}$       & $\mathbf{72.41}$        &  $\mathbf{97.47}$       &      $\mathbf{99.34}$   &    $\mathbf{77.18}$    &  $\mathbf{73.21}$     \\ [-0.5ex]\hline  
	\end{tabular}}}
\end{table*}

\subsection{Implementation Details}

The DMIC method is implemented on Pytorch platform, with two TITAN Xp GPU. We employ an effective network, i.e., AGW \cite{ye2021deep}, as our backbone network. AGW is designed on ResNet-50 \cite{he2016deep} with the differences that the first layer is replaced by two shallow modality-specific layers. Then, the backbone is initialized by the weights pretrained on ImageNet \cite{deng2009imagenet}. The whole training can be divided into intra-modality training for 50 epochs and inter-modality training for another 50 epochs. In the training phase, without any special emphasis, the batch size is set to 64, and $P$ and $K$ are set to 4 and 16, respectively. Several image augment methods, such as random horizontal flipping, random erasing, and random cropping, are employed in training. The images are resized to $288\times144$. The learning rate is $3.5\times10^{-4}$, and it is reduced by a factor of ten every 20 epochs. During the testing phase, the features after global average pooling layer are used to calculate the cosine similarity for evaluation. DBSCAN \cite{DBSCAN_Ester_1996} is employed to assign pseudo labels for unlabled data at the beginning of each epoch. For the SYSU-MM01 dataset, $\pi_2$, $\pi_1$, $\epsilon _3$, $\epsilon _2$, $\epsilon _1$, $\sigma_n$, $\sigma_b$, $\sigma_k$ and $k_1$ are set to 0.6, 0.3, 32, 18, 6, 0.5, 2, 3 and 40 respectively. For the RegDB dataset, these values are configured as 0.3, 0.2, 18, 12, 6, $\frac{2}{3}$, 1.5, 2 and 38, respectively. The temperature factor $\tau$ is set to 0.05. The trade-off weights $\varphi_1$ and $\varphi_2$ for losses are set to 0.1 and 1.0.

% \usepackage{multirow}
\begin{table*}[t]
	\centering
	\renewcommand{\multirowsetup}{\centering}
	\renewcommand{\arraystretch}{1.3}
		\caption{The performance(\%) comparison on RegDB dataset. \dag indicates training with 128 batch size.}\label{tab:comparison_regdb}
	\scalebox{0.8}
	{\setlength{\tabcolsep}{2.0 mm}{\begin{tabular}{c|cc|ccccc|ccccc}
				\hline
				& \multicolumn{2}{c|}{RegDB Settings} & \multicolumn{5}{c|}{Visible to Infrared}                 & \multicolumn{5}{c}{Infrared to Visible}               \\ [-1.0ex]
				& Methods                & Venue          & Rank-1 & Rank-10 & Rank-20 & mAP & mINP & Rank-1 & Rank-10 & Rank-20 & mAP & mINP \\ [-1.0ex] \hline\hline
				\multirow{9}{*}{\rotatebox{90}{Supervised}}
				& Zero-Padding \cite{wu2017rgb}             & ICCV-17        & 17.75   & 34.21    & 44.35  &18.90   & -        & 16.63     & 34.68      & 44.25       & 17.82       & -        \\ [-0.5ex]  
				& eBDTR \cite{ye2019bi}             & TIFS-19        & 34.62   & 58.96    & 68.72   &33.46    & -        & 34.21      & 58.74       & 68.64       & 32.49      & -        \\ [-0.5ex]     
				& X-Modal \cite{li2020infrared}               & AAAI-20        & 62.21   & 83.13    & 91.72    & 60.18    & -        & -      & -       & -       & -       & -        \\[-0.5ex]
				& Hi-CMD \cite{choi2020hi}              & CVPR-20        & 70.93   & 86.39    & -       & 66.04    & -        & -      & -       & -       & -       & -        \\[-0.5ex]
				& AGW \cite{ye2021deep}                   & TPAMI-21       & 70.05  & 86.21   & 91.55   & 66.37   & 50.19    & 70.49  & 87.21   & 91.84   & 65.90   & 51.24    \\[-0.5ex]
				& DDAG \cite{choi2020hi}                 & ECCV-20        & 69.34   & 86.19    & 91.49      & 63.46   &49.24       &68.06      & 85.15      & 90.31       & 61.80       & 48.62        \\ [-0.5ex]
				& VCD+VML \cite{tian2021farewell}               & CVPR-21        & 73.2  & -   & -   & 71.6   & -        & 71.8  & -   & -   & 70.1   & -        \\[-0.5ex]
				& CA \cite{ye2021channel}                    & ICCV-21        & 85.03  & 95.49   & 97.54  & 79.14   & 65.33    & 84.75  & 95.33  & 97.51   & 77.82   & 61.56    \\[-0.5ex]
				& MPANet \cite{ye2021channel}                    & CVPR-21        & 82.8  & -   & -  & 80.7   & -    & 83.7  & - & -   & 80.9   & -   \\[-0.5ex]
				& MSO \cite{gao2021mso}                   & MM-21          & 73.6  & 88.6   & -       & 66.9  & -        & 74.6  & 88.7   & -       & 67.5   & -        \\[-0.5ex]
				& MCLNet \cite{hao2021cross}                & ICCV-21        & 80.31  & 92.70   & 96.03   & 73.07   & 57.39    & 75.93  & 90.93   & 94.59   & 69.49   & 52.63    \\[-0.5ex]
				& SPOT \cite{chen2022structure}                & TIP-22        & 80.35  & 93.48  & 96.44   & 72.46   & 56.19    &79.37  & 92.79   & 96.01   & 72.26   & 56.06    \\[-0.5ex]
				& DMiR \cite{hu2022adversarial}                & TCSVT-22        & 75.79  & 89.86  & 94.18   & 69.97   & -    & 73.93  & 89.87  & 93.98  & 68.22  & -    \\[-0.5ex]
				& FMCNet  \cite{zhang2022fmcnet}               & CVPR-22        & 89.12  & -       & -       & 84.43   & -        & 88.38  & -       & -       & 83.86   & -        \\[-0.5ex]
				& MAUM \cite{liu2022learning}                   & CVPR-22        & 87.87  & -       & -       & 85.09   & -        & 86.95  & -       & -       & 84.34   & -        \\[-0.5ex] 				
				& TMD \cite{lu2023tri}                  & TMM-23        & 87.04  & 95.49      & 97.57       &81.19   & 68.73        &83.54  & 94.56      & 96.84       & 77.92  & 64.33       \\ [-0.5ex]\hline\hline
				\multirow{16}{*}{\rotatebox{90}{Unsupervised}} 
				
				& SSG  \cite{fu2019self}                 & ICCV-19        & 1.91  & 5.14   & 7.53   & 3.18   & -   &- & -   & -   & -   & -    \\
				& ECN  \cite{zhong2019invariance}                 & CVPR-19        & 2.17  &8.38   & 12.55  & 2.90  & -    & -  & -   & -   & -   & -   \\
				 & SPCL \cite{ge2020self} & NIPS-20  &13.59  & 26.98   & 34.88   & 14.86   & 10.36    &11.70  & 25.53   & 32.82   & 13.56   & 10.09    \\[-0.5ex]
 				& MMT  \cite{ge2020mutual}                  & ICLR-20        & 25.68  & 42.23   & 54.03   & 26.51   & 19.56   & 24.42 & 41.21   & 51.89   & 25.59  & 18.66   \\[-0.5ex]
				& CAP  \cite{wang2021camera}                   & AAAI-21        & 9.71 & 19.27    & 25.60   & 11.56   & 8.74     & 10.21  & 19.91   & 26.38   & 11.34   & 7.92    \\[-0.5ex]
				& IICS  \cite{xuan2021intra}                 & CVPR-21        & 14.39  & 47.91   & 62.32   & 15.74   & 8.41   & 15.91  & 54.20   & 71.49   & 24.87   & 22.15   \\		
				& ICE  \cite{chen2021ice}                 & ICCV-21        & 12.98  & 25.87  & 34.40   & 15.64  &11.91   & 12.18 & 25.67   & 34.90  & 14.82 & 10.60    \\

				& Cluster-Contrast \cite{dai2022cluster}      & ACCV-22         & 11.76  & 24.83   & 32.84    & 13.88  & 9.94    & 11.14  & 24.11   & 32.65   & 12.99   & 8.99    \\[-0.5ex]
				& PPLR  \cite{cho2022part}                 & CVPR-22        & 8.93  & 20.87   & 27.91   & 11.14   & 7.89     & 8.11  & 20.29   & 28.79  & 9.07   & 5.65    \\[-0.5ex]
				& ISE \cite{zhang2022implicit}                   & CVPR-22        & 16.12  & 23.30   & 28.93   & 16.99   & 13.24    & 10.83 & 18.64   &27.09   & 13.66   & 10.71    \\ [-0.5ex]
				 & H2H \cite{liang2021homogeneous}  & TIP-21         & 23.81 & 45.31   & 54.00   & 18.87  & -        & -      & -       & -       & -       & -        \\[-0.5ex]
				& OTLA \cite{wang2022optimal}                   & ECCV-22        & 32.90  & -   & -   & 29.70   & -        & 32.10   & -       & -       & 28.60    & -        \\[-0.5ex]
				& ADCA  \cite{yang2022augmented}                 & MM-22          & 67.20  & 82.02   & 87.44   & 64.05   &52.67    & 68.48  & 83.21   & 88.00   & 63.81   & 49.62    \\[-0.5ex]
				& CHCR \cite{pang2023cross}   & TCSVT-23   & 68.18      &  81.54       &   88.67     &   63.75     &   -      &   69.08     &  83.69       & 88.23      &  63.95     &  -        \\ [-0.5ex] 
				& TAA \cite{yang2023translation}   & TIP-23   & 62.23     & 80.00       &   86.02   &   56.00   &   41.51     & 63.79    & 81.80      & 86.80     &  56.53     &  38.99       \\ [-0.5ex] 
				& PGMAL \cite{wu2023unsupervised}   & CVPR-23   &69.48       & -      &   -      &   65.41      &   -      &   69.85     & -      &  -       &   65.17      &  -       \\ [-0.5ex]
				& CCLNet \cite{chen2023Unveiling}   & MM-23   &69.94       & -      &   -      &   65.53     &   -      &   70.17    & -      &  -       &  66.66     &  -       \\ [-0.5ex]
				& GUR \cite{yang2023Toward}   & ICCV-23   &73.91       & -      &   -      &  70.23      &   58.88      &   75.00    & -      &  -       &   69.94      &  56.21       \\ [-0.5ex]
				& DCCL \cite{yang2024Dual}   & TIFS-24   &78.28      & -      &   -      &   71.98    &  58.79     &   78.28    & -      &  -       &  71.30      &  55.23      \\ [-0.5ex] \hline\hline
				& $\mathbf{DMIC (ours)} $    & -       &   $\mathbf{86.31}$    &   $\mathbf{94.18}$  &    $\mathbf{96.31}$     &    $\mathbf{81.36}$    &   $\mathbf{70.31}$       &  $\mathbf{85.48}$   &  $\mathbf{93.47}$      &      $\mathbf{95.45}$   & $\mathbf{79.93}$      & $\mathbf{66.41}$     \\ [-0.5ex]
				& $\mathbf{DMIC^\dag (ours)}$   & -       & 83.97  & 92.67  & 95.00  & 78.54  & 66.44   & 83.79  & 92.40  & 94.62   & 77.65   & 63.23    \\ [-0.5ex]\hline  
	\end{tabular}}}

\end{table*}

\begin{table*}[h]
	\centering
	\renewcommand{\multirowsetup}{\centering}
	\renewcommand{\arraystretch}{1.5}
		\caption{Ablation studies for objective functions on SYSU-MM01 and RegDB (in $\%$).}\label{tab:Loss_Aalation_Study}
	\scalebox{0.8}
	{\setlength{\tabcolsep}{0.7mm}{\begin{tabular}{c|cccc|ccc|ccc|ccc}
				\hline
				& \multicolumn{4}{c|}{Components} & \multicolumn{3}{c|}{SYSU-MM01(All-search)} & \multicolumn{3}{c|}{SYSU-MM01(Indoor-search)} & \multicolumn{3}{c}{RegDB(Visible to Infrared)} \\
				Index & CC-1   & IC-1   & CC-2  & IC-2  & Rank-1           & Rank-10          & mAP          & Rank-1            & Rank-10           & mAP           & Rank-1             & Rank-10            & mAP                \\ \hline
				1& \ding{51}          &       &     &      & 32.47     & 72.91      & 30.6      & 37.94       & 79.92       & 46.86       & 31.17       & 48.98      & 32.78       \\
				2& \ding{51}          & \ding{51}     &     &      & 39.09     & 82.21     &38.08      & 47.79     & 86.89       & 56.14       & 48.88       & 65.97        & 47.54       \\
				3& \ding{51}          &   \ding{51}    & \ding{51}   &      &59.94      & 92.48      & 57.49     & 65.38      & 96.54       & 72.06     & 85.23      & 93.6       & 80.92      \\
				4& \ding{51}          & \ding{51}     & \ding{51}   &    \ding{51}  & 60.26      & 92.67      & 57.82     & 65.83       & 96.70       & 72.55      & 86.31       & 94.18        & 81.36       \\ \hline
	\end{tabular}}}

\end{table*}

\begin{table*}[h]
	\centering
	\renewcommand{\multirowsetup}{\centering}
	\renewcommand{\arraystretch}{1.5}
	\caption{Ablation studies for clustering on SYSU-MM01 and RegDB (in $\%$).}\label{tab:Clustering_Aalation Study}
	\scalebox{0.8}
	{\setlength{\tabcolsep}{0.7mm}{\begin{tabular}{c|cccc|ccc|ccc|ccc}
				\hline
				& \multicolumn{4}{c|}{Components} & \multicolumn{3}{c|}{SYSU-MM01(All-search)} & \multicolumn{3}{c|}{SYSU-MM01(Indoor-search)} & \multicolumn{3}{c}{RegDB(Visible to Infrared)} \\
				Index & VC   & MIE   & DNC & HMCL & Rank-1           & Rank-10          & mAP          & Rank-1            & Rank-10           & mAP           & Rank-1             & Rank-10            & mAP              \\ \hline
				1& \ding{51}          &       &     & \ding{51}       & 42.11    &82.41     &40.91      & 46.09     & 88.69      & 55.79     &  72.33      &  85.34     &  69.45         \\
				2& \ding{51}          &          &   \ding{51}  & \ding{51}   &43.44     & 82.47    &43.01      &47.83     & 89.44       &57.44      &  78.20       &  89.13      & 74.15      \\
				3&           &   \ding{51}       &   & \ding{51}   & 58.71     & 90.21     &55.99     & 62.45       & 93.32      & 68.95      & 84.49        & 92.71       & 79.93       \\
				4&          & \ding{51}       &    \ding{51} & \ding{51}  & 60.26      & 92.67      & 57.82     & 65.83       & 96.70       & 72.55       &86.31       & 94.18        & 81.36      \\ \hline
	\end{tabular}}}
	
\end{table*}

\begin{figure}
	\centerline{\includegraphics[width=0.5\textwidth]{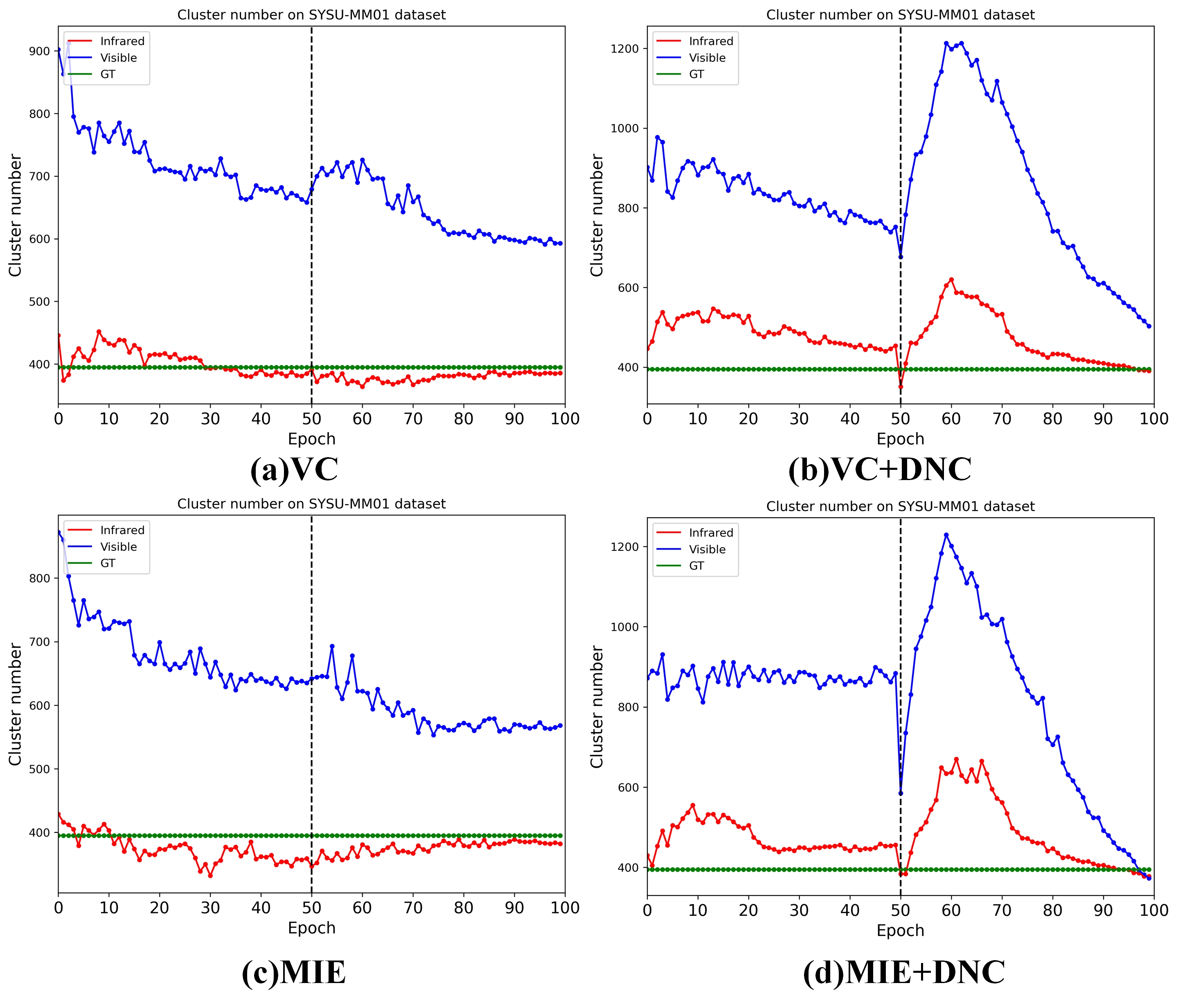}}
	\caption{The cluster numbers on the SYSU-MM01 dataset are displayed using the following experimental configurations: (a) VC, (b) VC+DNC, (c) MIE, and (d) MIE+DNC. The number of infrared clusters is represented by the red line, the number of visible clusters is depicted by the blue line, and the number of ground truth clusters is represented by the green line.}
	\label{fig:cluster_ablation}       % Give a unique label
\end{figure}

\begin{figure}
	\centerline{\includegraphics[width=0.3\textwidth]{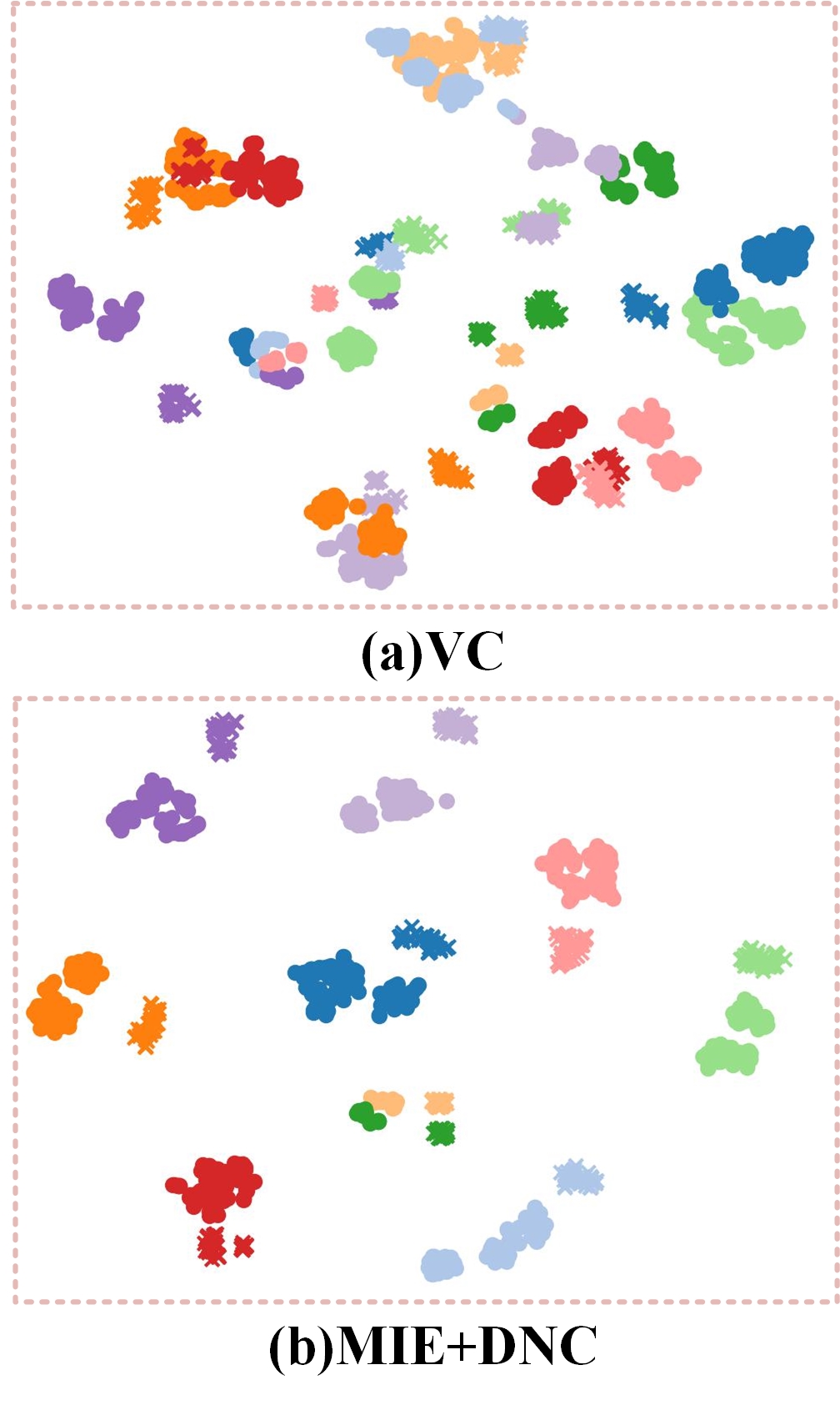}}
	\caption{The visualization showcases the distribution of 10 randomly chosen identities from the SYSU-MM01 dataset. The experimental results depicted are obtained from two different configurations: (a) VC and (b) MIE+DNC. Different colors represent different identities. The 'o' represents visible samples and the 'x' represents infrared samples.}
	\label{fig:distributions}       % Give a unique label
\end{figure}

\subsection{Comparison with State-of-the-art Methods}
In this subsection, we perform a comparative analysis, evaluating our unsupervised DMIC method against 16 state-of-the-art supervised methods and 19 state-of-the-art unsupervised methods. The goal is to highlight the competitive performance of our approach. Due to computational resource limitations, the majority of our experiments are conducted with a batch size of 64. However, we find that training with a batch size of 128 significantly improves model performance. Therefore, we present the performance achieved with batch sizes of 64 and 128. This is much smaller than the 256 batch size used in recent methods, making it more economical in terms of training.

\textbf{Performance Comparison on SYSU-MM01 dataset.}\quad Table \ref{tab:comparison_sysu} presents the performance comparison of our unsupervised method with state-of-the-art methods on the SYSU-MM01 dataset. Firstly, we present the performance of selected methods that have demonstrated promising results in the field of unsupervised person ReID (USL-ReID) when applied to the task of unsupervised visible-infrared person ReID (USL-VI-ReID). These methods include SSG \cite{fu2019self}, ECN \cite{zhong2019invariance}, SPCL \cite{ge2020self}, MMT \cite{ge2020mutual}, CAP \cite{wang2021camera}, IICS \cite{xuan2021intra}, ICE \cite{chen2021ice}, Cluster-Contrast \cite{dai2022cluster}, PPLR \cite{cho2022part}, and ISE \cite{zhang2022implicit}. In particular, ICS, CAP and ICE also take into account the elimination of camera differences. To be specific, we outperform ICE by $44.7\%$/$41.62\%$ in Rank-1 and  $37.43\%$/$34.2\%$ in mAP under two testing modes. These results clearly illustrate that these methods designed for USL-ReID do not seamlessly apply to the USL-VI-ReID task, even when accounting for camera-related factors. They fail to address modal discrepancy, leading to poor performance on cross-modal dataset. Secondly, we achieve SOTA performance when compared to recent USL-VI-ReID, including H2H \cite{liang2021homogeneous}, OTLA \cite{wang2022optimal}, ADCA \cite{yang2022augmented}, CHCR \cite{pang2023cross}, TAA \cite{yang2023translation}, PGMAL \cite{wu2023unsupervised}, CCLNet \cite{chen2023Unveiling}, GUR \cite{yang2023Toward}, and DCCL \cite{yang2024Dual}. Although we do not pre-train the model on labeled single modality person dataset like the H2H and TAA methods to facilitate the assignment of pseudo labels, we are still able to establish robust cross-modal associations and achieve promising results. We outperform PGMAL by a large margin of $2.99\%$/$9.6\%$ in Rank-1 and $6.04\%$/$9.83\%$ in mAP under both testing modes. Compared with GUR method, we also achieve a considerable improvement by $2.06\%$/$1.3\%$ in Rank-1 and $0.38\%$/$0.95\%$ in mAP under both testing modes. Finally, we achieve performance on par with MCLNet \cite{hao2021cross} and SPOT \cite{chen2022structure} and even surpass the performance of several supervised methods, including Zero-Padding \cite{wu2017rgb}, eBDTR \cite{ye2019bi}, X-Modal \cite{li2020infrared}, Hi-CMD \cite{choi2020hi}, AGW \cite{ye2021deep}, DDAG \cite{choi2020hi}, VCD+VML \cite{tian2021farewell}, MSO \cite{gao2021mso}, DMiR \cite{hu2022adversarial} and FMCNet \cite{zhang2022fmcnet}. These results suggest that unsupervised methods hold promise for achieving performance close to that of supervised methods.

\textbf{Performance Comparison on RegDB dataset.}\quad We compare our unsupervised DMIC method with several impressive supervised and unsupervised methods on RegDB dataset. We present the performance achieved with batch sizes of 64 and 128. 
The experimental results indicate that the optimal results can be achieved through training with a batch size of 64. Compared to the unsupervised method, we significantly improve the upper bound of SOTA performance from $78.28\%$/$78.28\%$ to $86.31\%$/$85.48\%$ in Rank-1 and $71.98\%$/$71.30\%$ to $81.36\%$/$79.93\%$ in mAP. These enhancements are attributed to our meticulously designed clustering method and objective functions. Compared to recent supervised methods, we achieve comparable performance. This shows that on the RegDB dataset, our method holds promise in reducing dependence on manual labeling. Training without relying on labels can also achieve performance comparable to supervised methods.

In this subsection, we compare our unsupervised DMIC method with the state-of-the-art methods on two mainstream databases. The batch size in recent works \cite{yang2022augmented, wu2023unsupervised} is $16\times16$, necessitating a total of 64GB of GPU memory. Our approach uses much less GPU memory, with a $4\times16$ batch size requiring only 22G of GPU memory to achieve competitive performance. The training overhead is greatly reduced. The competitive performance of our method can be attributed to three major advantages: 1) Our approach effectively handles both cross-modality and cross-camera discrepancies at the clustering level, thereby bolstering the accuracy of label estimation and the robustness of the optimization objective. 2) Adaptation of clustering objectives dynamically steers the network's learning process from simplicity to complexity. Prioritizing reliable samples initially amplifies discriminative capabilities, while subsequently integrating cross-camera and cross-modal samples enhances generalization skills. Notably, our clustering method can be easily integrated into any unsupervised framework. 3) Cluster-level and instance-level objective functions can facilitate network learning to modality-invariant feature representations. Overall, our method enhances the performance of USL-VI-ReID, rendering it a promising solution for real-world applications.

\begin{figure}
	\centerline{\includegraphics[width=0.5\textwidth]{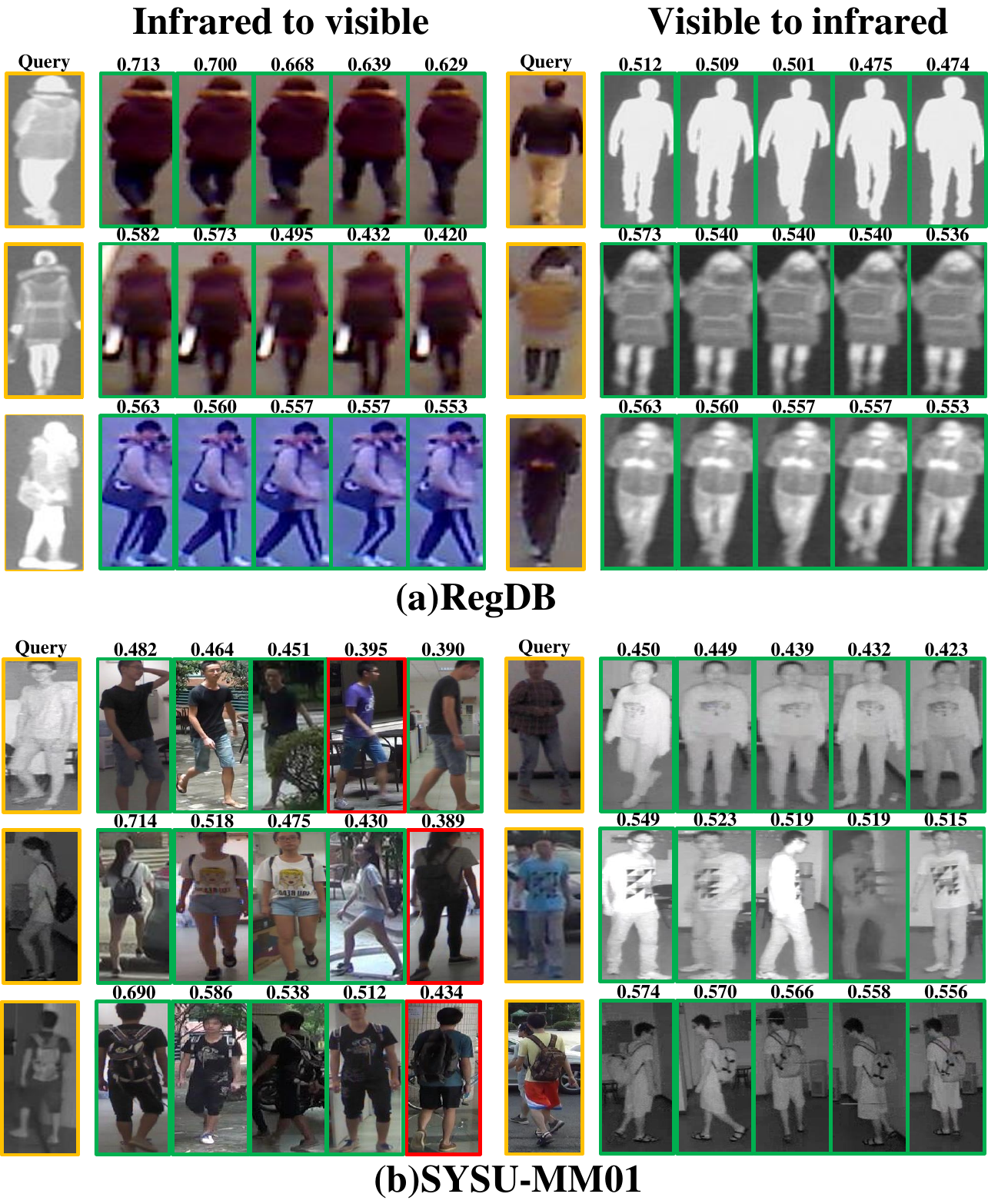}}
	\caption{The top-5 retrieved results on (a) RegDB and (b) SYSU-MM01 dataset are presented. Correct matchings are identified by green bounding boxes, wrong matchings by red bounding boxes, and query samples by yellow bounding boxes. The cosine similarity values are shown above the gallery images.}
	\label{fig:visual_pic}       % Give a unique label
\end{figure}

\subsection{Various Properties Analysis of the DMIC Model}

In this subsection, we verify the effectiveness of each key module in our DMIC method on SYSU-MM01 and RegDB datasets. Ablation studies on objective functions and clustering strategy are conducted to exhaustively analyze the performance of the network. 

\textbf{Ablation study for objective functions.}\quad As listed in Table \ref{tab:Loss_Aalation_Study}, we analyze the contribution of each objective function. CC-1 and IC-1 represent the cluster-level and instance-level losses during intra-modality training, while CC-2 and IC-2 refer to the cluster-level and instance-level losses during inter-modality training. When gradually superimposing the objective function IC-1, CC-2 and IC-2 on CC-1, the performance of the model can be effectively improved. The largest improvement lies in the introduction of CC-2, which nearly doubles the model performance on the Regdb dataset. CC-2 utilizes robust pseudo-labels assigned by the MIE and DNC, which are modality and camera-independent, to facilitate the learning of modality-camera invariant representations. Additionally, IC-1 and IC-2 effectively refine the relationships between instances, which leads to performance improvements.

\textbf{Ablation study for clustering.}\quad  Clustering is widely recognized as a critical component in unsupervised learning frameworks. As shown in Table \ref{tab:Clustering_Aalation Study}, we analyze the effectiveness of MIE and DNC. VC refers to the vanilla clustering method employed in recent studies \cite{yang2022augmented,wu2023unsupervised,yang2023translation}. Results from experiment index 3 reveal a substantial improvement of approximately 10\%-15\% in mAP with the proposed MIE. By incorporating our MIE module, our unsupervised framework is able to generate more robust pseudo labels, leading to a significant enhancement in performance. From index 2 and 4, we can oberve that DNC can boost the performance of 2\%-5\% in mAP. The dynamic strategy of DNC enables further refinement of the network's optimization objective through clustering. By progressively transitioning from hard to easy objectives, we are able to systematically elevate the upper bound of the model's performance. Overall, the enhancements in clustering achieved by our approach contribute to a significant boost in performance, with improvements ranging from 14\%-17\% in mAP. This clearly illustrates the effectiveness of our association approach in handling cross-modal and cross-camera samples, which in turn offers fresh insights into the USL-VI-ReID task.

\textbf{Visualization Analysis.}\quad To further substantiate the efficacy of our method, we perform visualization experiments. Fig. \ref{fig:cluster_ablation} illustrates the cluster number evolution during training.  Fig. \ref{fig:distributions} shows the distribution \cite{t-sne} of 10 randomly selected challenging identities from the SYSU-MM01 dataset. As displayed in Fig. \ref{fig:cluster_ablation}, the cluster number of visible samples is prone to suffer from identity splitting problem. This is primarily attributed to the variations in color and lighting present in visible images, caused by disparities in RGB camera characteristics. The number of visible clusters in the VC method decreases to approximately 600 during intra-modality and inter-modality training. Introducing the MIE module helps alleviate the identity splitting problem, resulting in a combined number of visible clusters of around 550. However, the lack of refined clustering objectives limits the cross-modal and cross-camera generalizability, making cluster merging difficult. Fig. \ref{fig:cluster_ablation} (d) illustrates that the collaboration between MIE and DNC is effective. This collaboration enhances the cross-modal and cross-camera generalizability of the model. It brings the number of visible and infrared clusters closer to the number of ground truth clusters. To be specific, we narrow down the $eps$ from $\pi_2$ to $\pi_1$ to prioritize discriminability learning from reliable clusters in the first 50 intra-modality learning epochs. This process is repeated for an additional 10 inter-modality learning epochs to further split solid single-camera clusters. During the last 40 inter-modality learning epochs, we widen $eps$ from $\pi_1$ to $\pi_2$ to merge cross-modal and cross-camera instances into clusters, thereby facilitating the overall cross-modal and cross-camera generalizability of the model. As seen in Fig. \ref{fig:distributions} (a), we can observe that using VC results in a substantial separation of identities within nearly all clusters. This separation hampers the model's classification capabilities and introduces noise, which adversely affect the model's performance. However, by incorporating MIE and DNC (see Fig. \ref{fig:distributions} (b)), we can learn compact representations for each cluster.

\textbf{Retrieved Results.}\quad As illustrated in Fig. \ref{fig:visual_pic}, we display retrieved results on SYSU-MM01 and RegDB datasets under infrared to visible and visible to infrared modes. The results demonstrate that DMIC is effective in learning modality-camera invariant features. Even in the presence of challenging and confusing samples, it is still possible to maintain a relatively high similarity among positive samples.

\section{Conclusion}
This paper presents a novel Dynamic Modality-Camera Invariant Clustering (DMIC) method for USL-VI-ReID task. Specifically, Modality-Camera Invariant Expansion (MIE) integrates inter-modal and inter-camera distance coding, which effectively bridges cross-modality and cross-camera gaps at clustering-level. Dynamic Neighborhood Clustering (DNC) employs two dynamic search strategies. The dynamic search strategies help to refine the network's optimization objective through clustering, from discriminability ability enhancement to cross-modal and cross-camera generalizability learning. Moreover, a Hybrid Modality Contrastive Learning (HMCL) is designed to optimize instance-level and cluster-level distributions. The memories for intra-modality and inter-modality training are updated by randomly selected samples, which help to fully explore modality-invariant representations in a real-time manner. Extensive experiments have demonstrated that our DMIC addresses the limitations present in current clustering approaches and significantly perform better than current state-of-the-art unsupervised methods. These findings provide valuable insights for the USL-VI-ReID task and further reduce the performance gap with supervised methods.

\end{document}